\documentclass[11pt,draftcls,onecolumn]{IEEEtran}%
\usepackage{epsfig,epsf,psfrag}
\usepackage{amssymb,amsmath,amsthm,amsfonts,latexsym}
\usepackage{amsmath,graphicx,bm,xcolor,url}
\usepackage{algorithm}
\usepackage{epstopdf}
\catcode`~=11 \def\UrlSpecials{\do\~{\kern -.15em\lower .7ex\hbox{~}\kern .04em}} \catcode`~=13 

\allowdisplaybreaks[3]

\newcommand{\calI}{\mathcal{I}}

\newcommand{\calM}{\mathcal{M}}
\newcommand{\calN}{\mathcal{N}}
\newcommand{\calO}{\mathcal{O}}

\newcommand{\calU}{\mathcal{U}}

\newcommand{\ba}{\mathbf{a}}
\newcommand{\bA}{\mathbf{A}}

\newcommand{\bB}{\mathbf{B}}

\newcommand{\bD}{\mathbf{D}}
\newcommand{\be}{\mathbf{e}}
\newcommand{\bE}{\mathbf{E}}

\newcommand{\bh}{\mathbf{h}}

\newcommand{\bI}{\mathbf{I}}

\newcommand{\bn}{\mathbf{n}}
\newcommand{\bN}{\mathbf{N}}

\newcommand{\bQ}{\mathbf{Q}}

\newcommand{\bs}{\mathbf{s}}
\newcommand{\bS}{\mathbf{S}}
\newcommand{\bt}{\mathbf{t}}

\newcommand{\bu}{\mathbf{u}}
\newcommand{\bU}{\mathbf{U}}

\newcommand{\bx}{\mathbf{x}}
\newcommand{\bX}{\mathbf{X}}

\newcommand{\bY}{\mathbf{Y}}

\DeclareMathAlphabet{\mathbsf}{OT1}{cmss}{bx}{n}
\DeclareMathAlphabet{\mathssf}{OT1}{cmss}{m}{sl}% slanted sans serif

\DeclareSymbolFont{bsfletters}{OT1}{cmss}{bx}{n}  
\DeclareSymbolFont{ssfletters}{OT1}{cmss}{m}{n}
\DeclareMathSymbol{\bsfGamma}{0}{bsfletters}{'000}
\DeclareMathSymbol{\ssfGamma}{0}{ssfletters}{'000}
\DeclareMathSymbol{\bsfDelta}{0}{bsfletters}{'001}
\DeclareMathSymbol{\ssfDelta}{0}{ssfletters}{'001}
\DeclareMathSymbol{\bsfTheta}{0}{bsfletters}{'002}
\DeclareMathSymbol{\ssfTheta}{0}{ssfletters}{'002}
\DeclareMathSymbol{\bsfLambda}{0}{bsfletters}{'003}
\DeclareMathSymbol{\ssfLambda}{0}{ssfletters}{'003}
\DeclareMathSymbol{\bsfXi}{0}{bsfletters}{'004}
\DeclareMathSymbol{\ssfXi}{0}{ssfletters}{'004}
\DeclareMathSymbol{\bsfPi}{0}{bsfletters}{'005}
\DeclareMathSymbol{\ssfPi}{0}{ssfletters}{'005}
\DeclareMathSymbol{\bsfSigma}{0}{bsfletters}{'006}
\DeclareMathSymbol{\ssfSigma}{0}{ssfletters}{'006}
\DeclareMathSymbol{\bsfUpsilon}{0}{bsfletters}{'007}
\DeclareMathSymbol{\ssfUpsilon}{0}{ssfletters}{'007}
\DeclareMathSymbol{\bsfPhi}{0}{bsfletters}{'010}
\DeclareMathSymbol{\ssfPhi}{0}{ssfletters}{'010}
\DeclareMathSymbol{\bsfPsi}{0}{bsfletters}{'011}
\DeclareMathSymbol{\ssfPsi}{0}{ssfletters}{'011}
\DeclareMathSymbol{\bsfOmega}{0}{bsfletters}{'012}
\DeclareMathSymbol{\ssfOmega}{0}{ssfletters}{'012}

\newtheorem{theorem}{Theorem} 
\newtheorem{lemma}[theorem]{Lemma}

\newtheorem{corollary}[theorem]{Corollary}
\newtheorem{definition}{Definition}

\newtheorem{remark}{Remark}

\newtheorem{data model}{Data Model}

\begin{document}
\title{Robust Projection based Anomaly Extraction (RPE) in Univariate Time-Series }

\markboth{}%
%\markboth{Journal of \LaTeX\ Class Files,~Vol.~11, No.~4, %December~2012}%
{Shell \MakeLowercase{\textit{et al.}}: Bare Demo of IEEEtran.cls for Journals}
\author{Mostafa Rahmani, Anoop Deoras, Laurent Callot \\
mostrahm@amazon.com
}
% make the title area
\maketitle

\begin{abstract}
This paper presents a novel, closed-form, and data/computation efficient online anomaly detection algorithm for time-series data. The proposed method, dubbed RPE, is a window-based method and in sharp contrast to the existing window-based methods, it is robust to the presence of anomalies in its window and it can distinguish the anomalies in time-stamp level. RPE leverages the linear structure of the trajectory matrix of the time-series and employs a robust projection step which makes the algorithm able to handle the presence of multiple arbitrarily large anomalies in its window. A closed-form/non-iterative algorithm for the robust projection step is provided and it is proved that it can identify the corrupted time-stamps. RPE is a great candidate for the applications where a large training data is not available which is the common scenario in the area of time-series. An extensive set of numerical experiments show that RPE can  outperform  the existing approaches with a notable margin.
\end{abstract}

\section{Introduction}
Anomaly detection is an important research problem in unsupervised learning where the main task is to identify the observations which do not follow the common structure of the data. Anomalies  mostly correspond to rare but important events whose detection is of  paramount importance. For instance, in  medical imaging, the outlying patches could correspond to  malignant tissues \cite{karrila2011comparison}, and in computer networks  an anomaly  can imply an intrusion
attack \cite{kruegel2003anomaly}. 

In many applications and businesses, the given data is a time-series of real values where each value represents the value of an important metric/sensor/measurement at a time-stamp and the values are mostly sampled with a fixed frequency \cite{braei2020anomaly}.  In these applications, it is important to identify if (based on the past observation) a given  time-stamp value  is an anomaly or if it is a part of an anomalous pattern. An anomaly detection algorithm for time-series data is preferred to have two desirable properties: 
\\
\textbf{(a)} The algorithm should be able to declare an anomaly without any delay. In other word, the algorithm  only relies on the past observations. \\
\textbf{(b)} If an anomaly happens in time-stamp $t$, the algorithm should only label time-stamp $t$ as anomalous and the anomaly label should not be diffused to the neighboring time-stamps. We will discuss this important property further in the next sections since most window-based  methods \cite{braei2020anomaly,guha2016robust,ren2019time} can keep reporting anomalies for a long period of time after the anomaly occurred, as long as their window contains the anomaly.

This paper presents a new window-based algorithm which, in sharp contrast to the current window-based methods, is provably robust to the presence of anomalies in its window. The algorithm, dubbed RPE, can distinguish outlying  behaviors at the time-stamp level. 
The main contributions of this work can be summarized as follows. 
\\
$\bullet$ We present an accurate and data/computationally efficient algorithm which is based on transforming the problem of anomaly detection in uni-variate time series into the problem of robust projection into a linear manifold.
\\
$\bullet$ To the best of our knowledge, RPE is the only window-based method which is provably robust to the presence of outlying time-stamps in its window. 
\\
$\bullet$ 
The projection step involves solving a convex optimization problem. 
A closed form method is proposed which saves the algorithm from running an iterative solver for each time-stamp. 
\\
$\bullet$ Novel theoretical results are established which guarantee the performance of the  projection step. 
\\
$\bullet$ An extensive set of  experiments with real/synthetic data show that RPE can notably outperform the former approaches.

\subsection{Definitions and notation}
Bold-face upper-case and lower-case letters are used to
denote matrices and vectors, respectively. For a vector $\ba$, $\| \ba \|_p$ denotes its $\ell_p$-norm, $\ba(i)$ denotes its $i^{\text{th}}$ element, and $|\ba|$ is a vector whose values are equal to the absolute value of the corresponding elements in $\ba$. Similarly, the elements of $\bY = |\bX|$ are equal to the absolute value of the elements of matrix $\bX$. 
For a matrix $\bA$, $\bA^T$ is the transpose of $\bA$ and $\ba_i$ indicates the $i^{th}$ row of $\bA$.
$\mathbb{S}^{M_1 -1}$ indicates the unit $\ell_2$-norm sphere in $\mathbb{R}^{M_1}$.
Subspace $\calU^{\perp}$ is  the complement of $\calU$. For a set $\calI$, $|\calI|$ indicates the cardinality of $\calI$.

The coherency of a subspace $\calU$ with the standard basis is  a measure of the sparsity of  the vectors which lie in $\calU$ \cite{chandrasekaran2011rank,candes2009exact}. The following definition provides three metrics which can represent the coherency of a linear subspace. 
\begin{definition}
Suppose $\bU \in \mathbb{R}^{M_1 \times r}$ is an orthonormal matrix. Then we define
 \begin{eqnarray}
  \begin{aligned}
& \mu^2(\bU) = \max_i \frac{\| \be_i ^T \bU \|_2^2}{r}
\:,
\\
& \gamma(\bU) = \frac{1}{\min_{\bh \in \mathbb{S}^{r-1}} \| \bU \:\bh \|_1} \:,
\\
& \kappa (\bU) = {\mu (\bU)}{\gamma(\bU)} \:,
  \end{aligned}
  \label{eq:incoh}
\end{eqnarray}
where $\be_i$ is the $i^{th}$ row of the identity matrix.
\end{definition}

\noindent
The parameter $\mu^2(\bU)$ shows how close  $\calU$ (the column space of $\bU$) is to the standard basis and $\gamma(\bU) $ measures the minimum of the $\ell_1$-norm of all the vectors on the intersection of $\mathbb{S}^{r-1}$ and $\calU$. Clearly, the more sparse are the vectors in $\calU$, the larger are $\mu^2(\bU)$ and $\gamma(\bU) $. 

Suppose we have a set of samples from a random variable saved in set $\calM$. The empirical   CDF value corresponding to a sample $\bx$ is defined as $\frac{|\calM < x|}{|\calM|}$ where $| \calM < x|$ is the number of elements in $\calM$ which are smaller than $x$ and $|\calM|$ is the size of $\calM$. The trajectory matrix $\bX \in \mathbb{R}^{M_1 \times (n - M_1 + 1)}$ corresponding to time-series $\bt \in \mathbb{R}^n$ is defined as
$$
\bX = 
\begin{bmatrix}
\bt(1) & \bt(2) & \text{...} & \bt(n - M_1 + 1)\\
\bt(2) & \bt(3) & \text{...} & \bt(n - M_1 + 2)
\\
\vdots &\vdots &   & \vdots \\
\bt(M_1) & \bt(M_1 + 1) & \text{...} & \bt(n)
\end{bmatrix} \:,
$$
where $M_1$ is the size of the running window and define $M_2=n - M_1 + 1$ ($\bX$ is created by running a window of size $M_1$ on $\bt$).

\begin{figure*}%[h!]
\begin{center}
\mbox{
\includegraphics[width=1.34in]{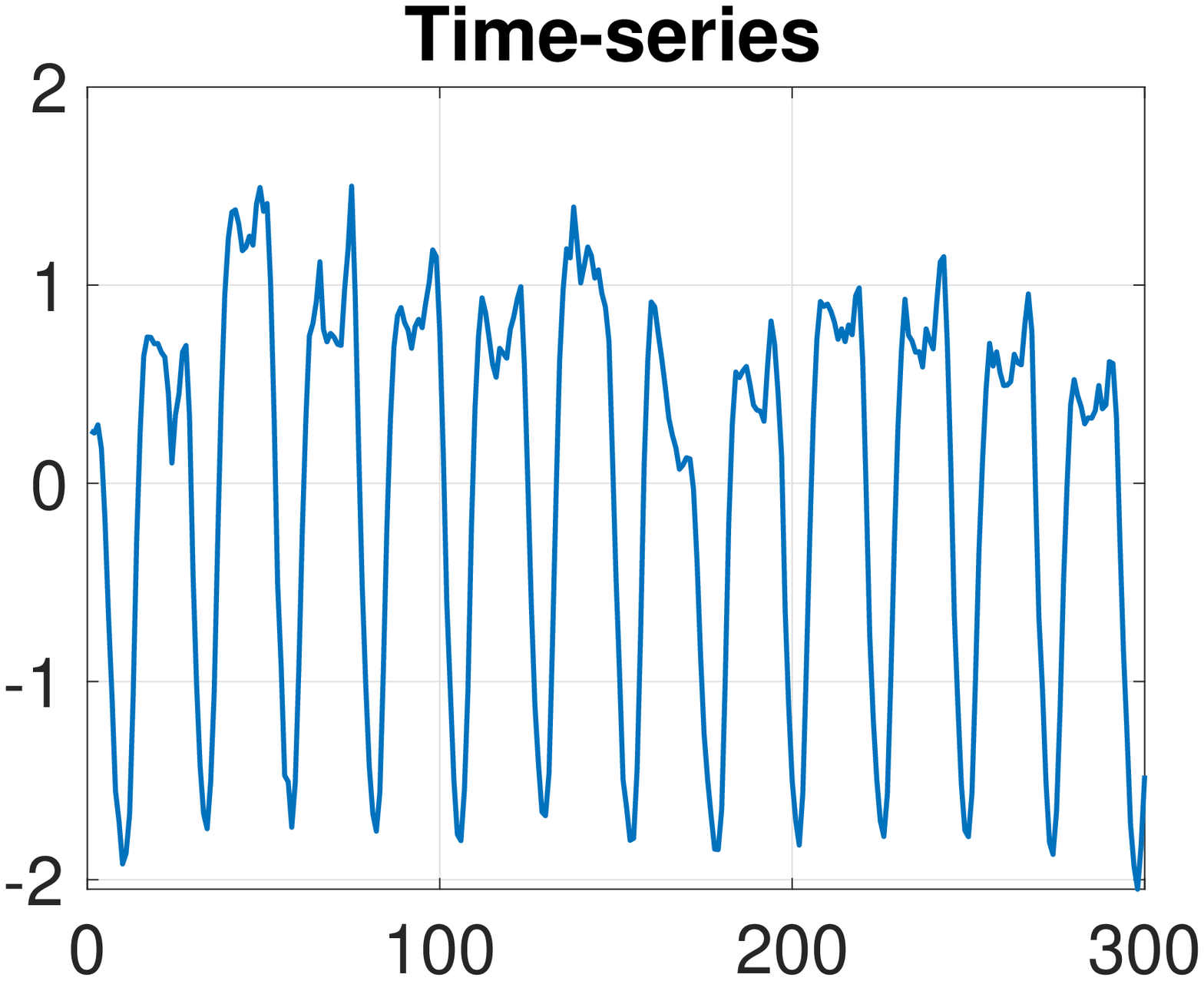}
\includegraphics[width=1.34in]{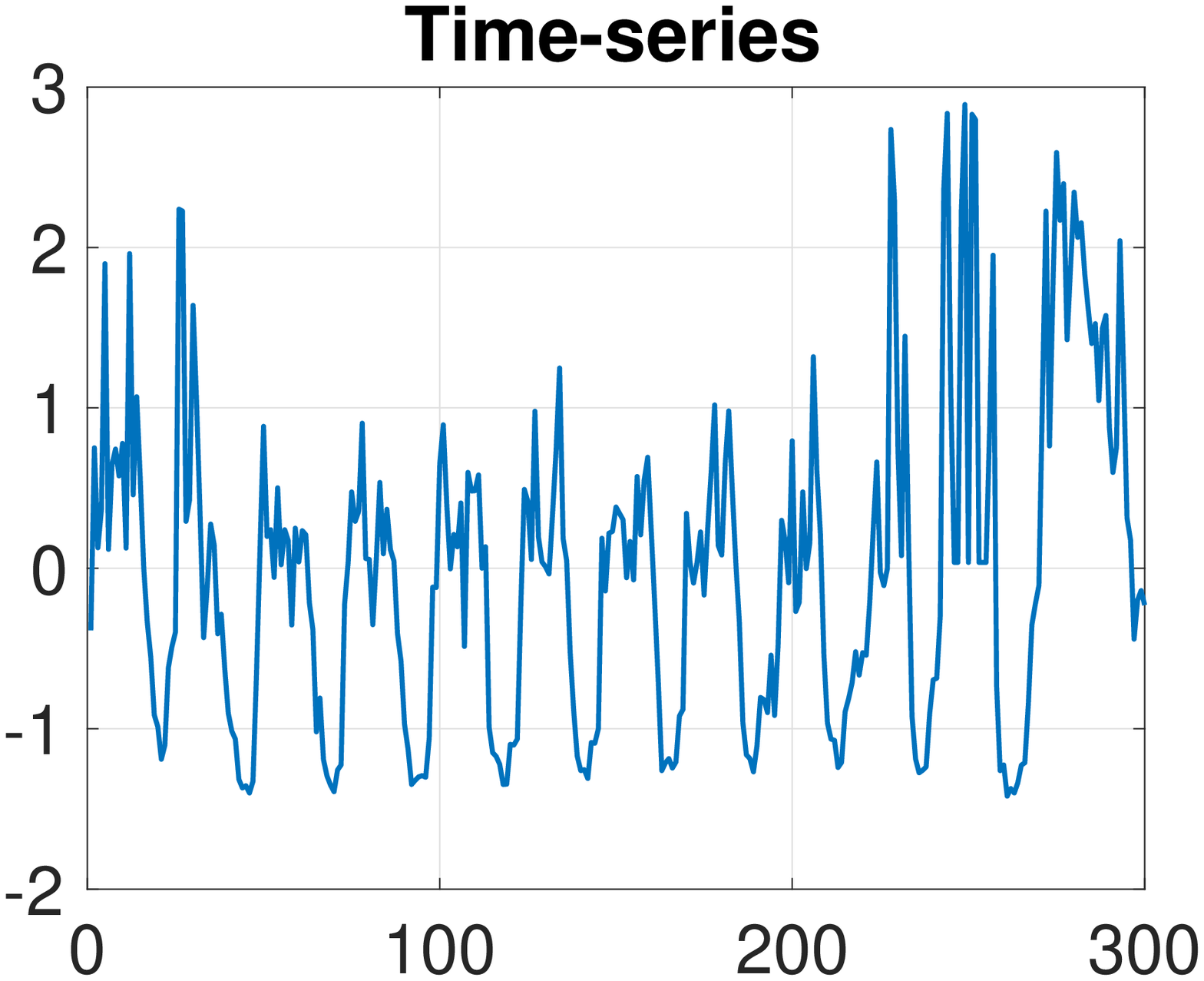}
\includegraphics[width=1.34in]{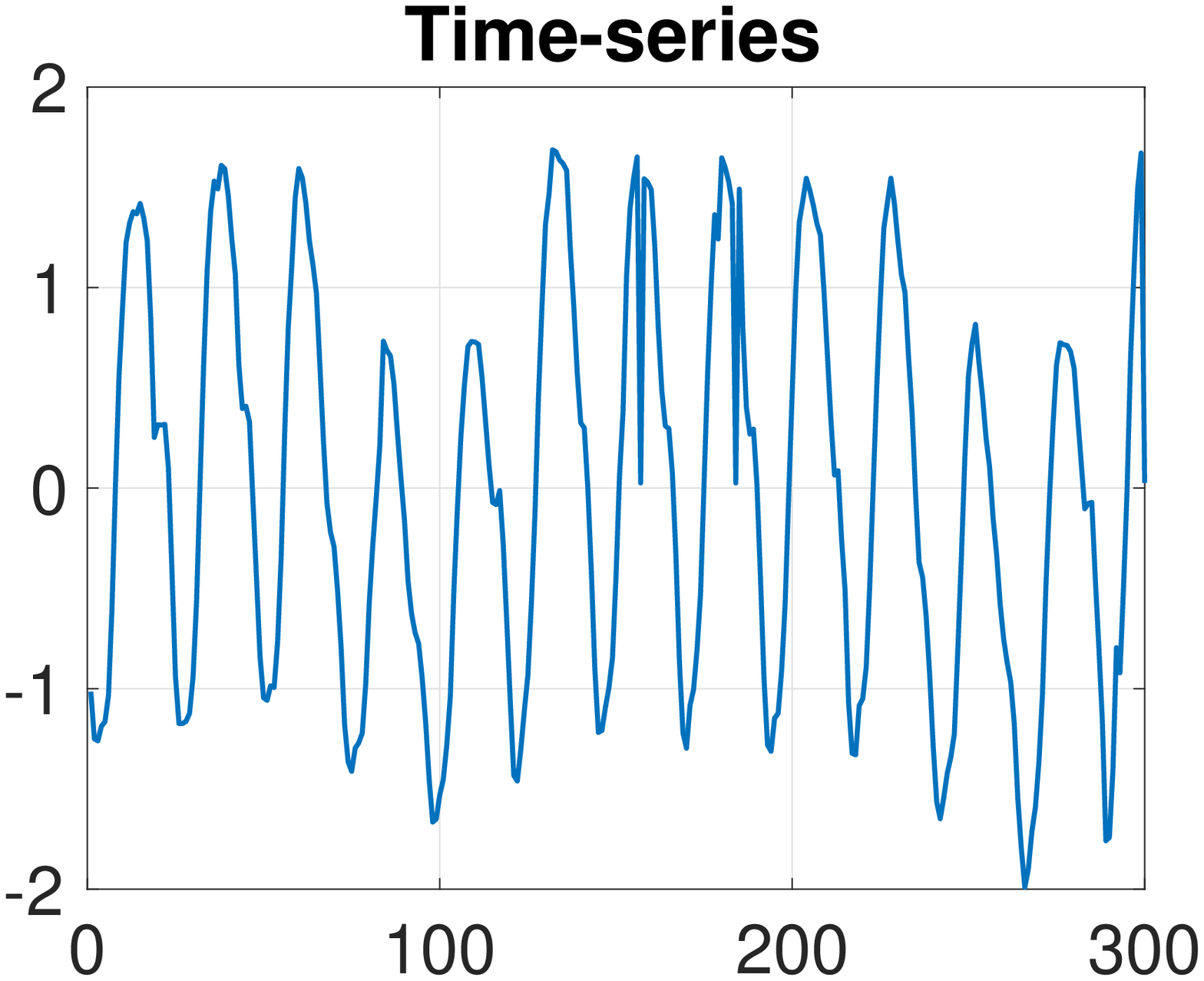}
\includegraphics[width=1.34in]{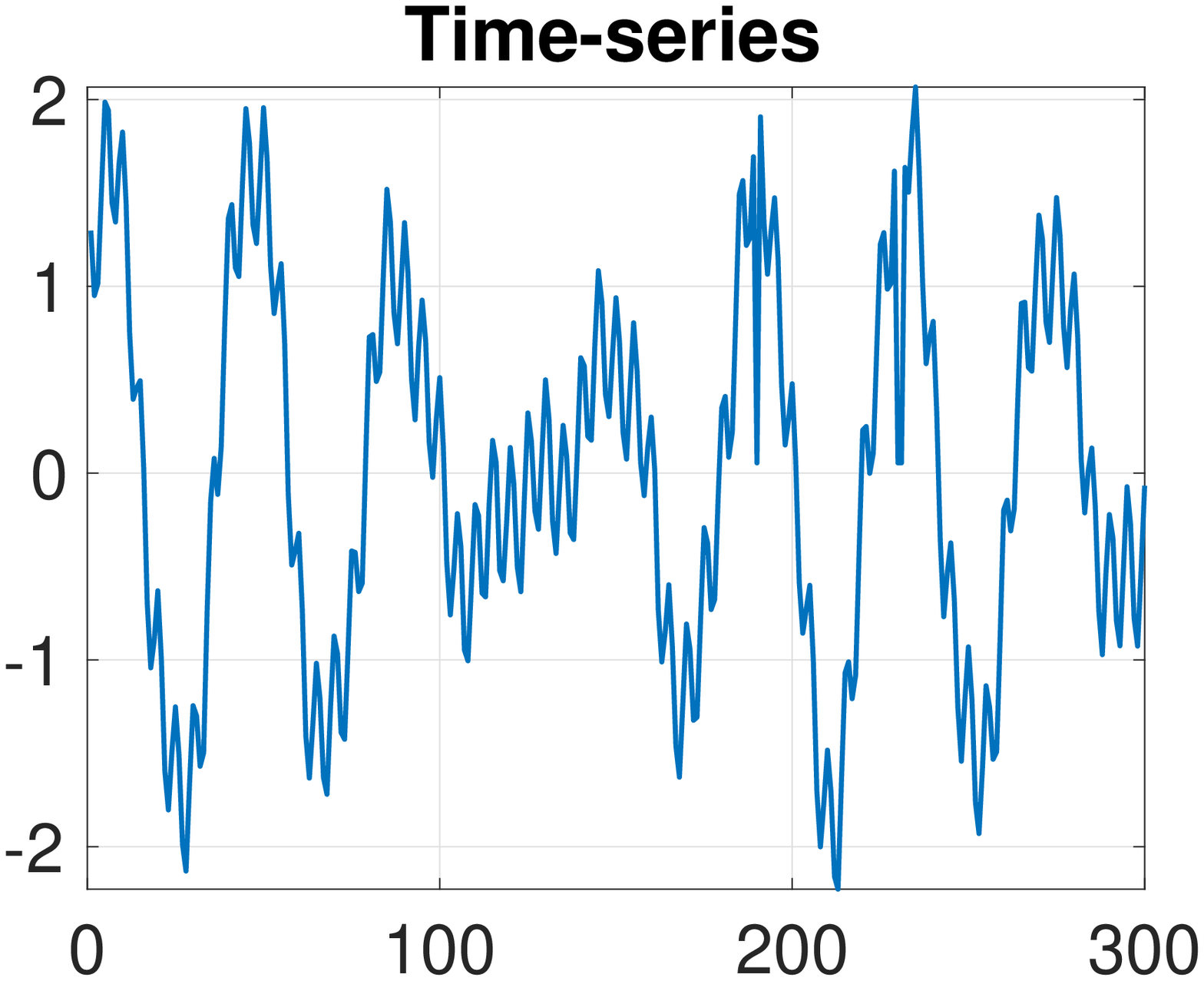}
\includegraphics[width=1.34in]{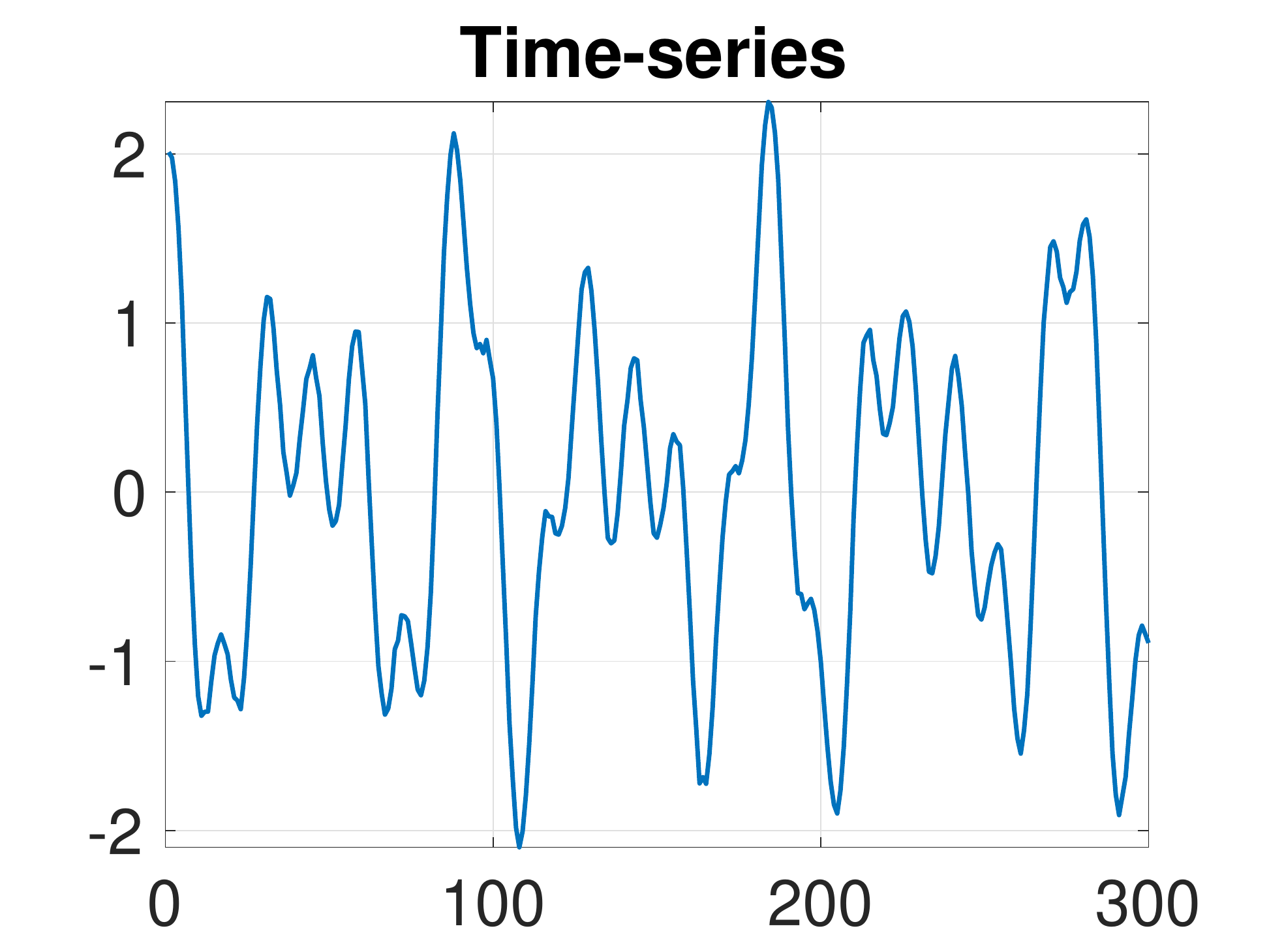}
}
\mbox{
\includegraphics[width=1.34in]{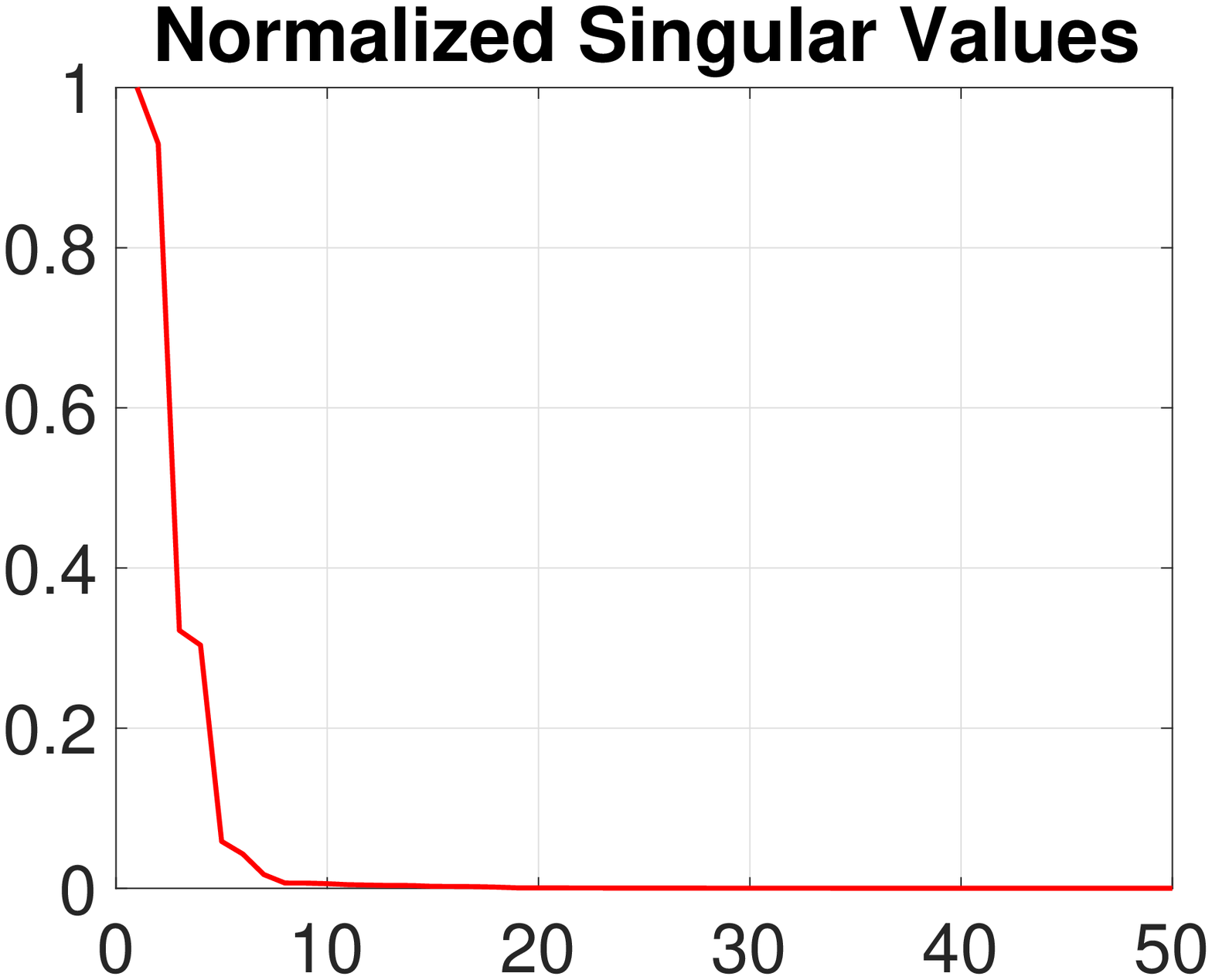}
\includegraphics[width=1.34in]{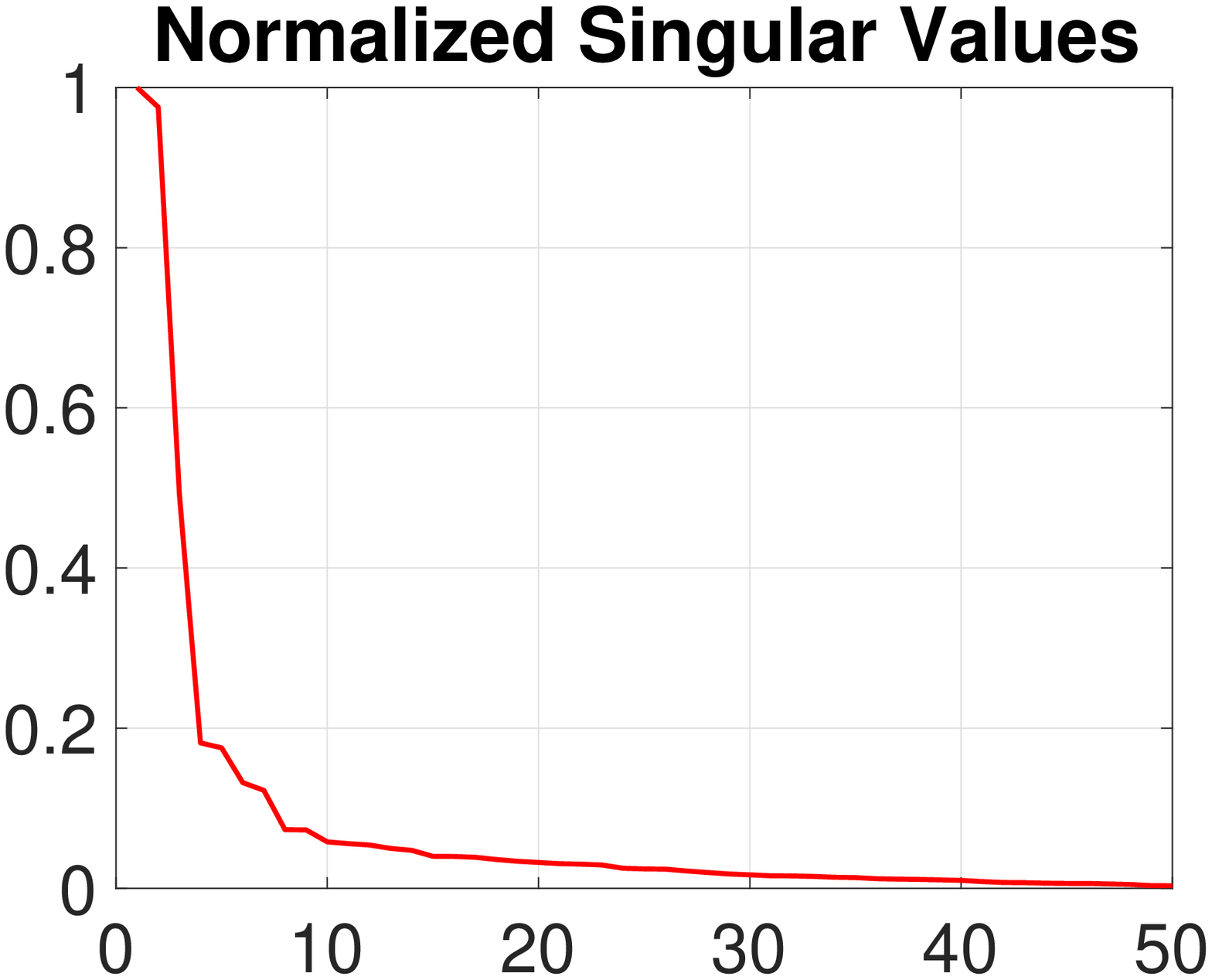}
\includegraphics[width=1.34in]{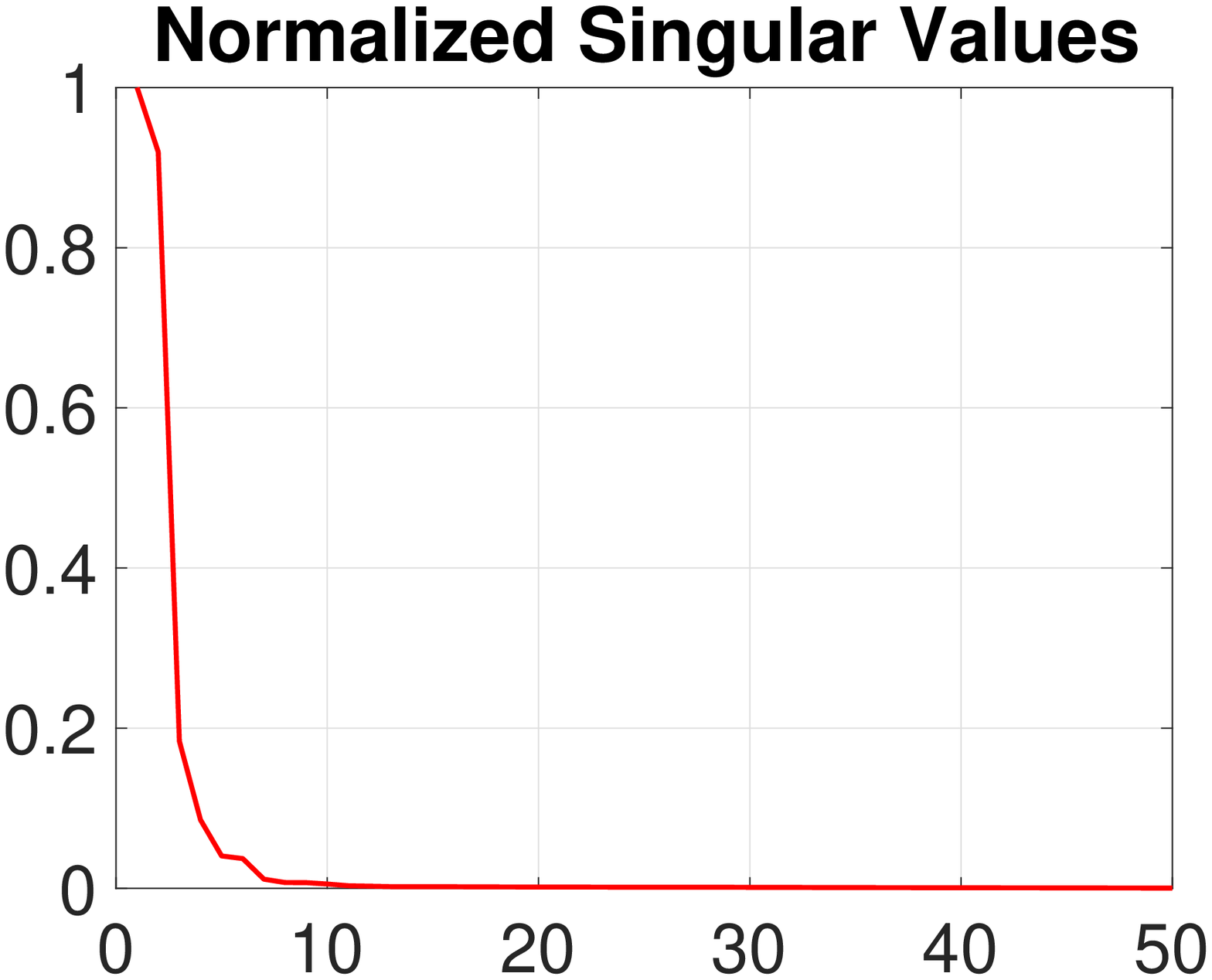}
\includegraphics[width=1.34in]{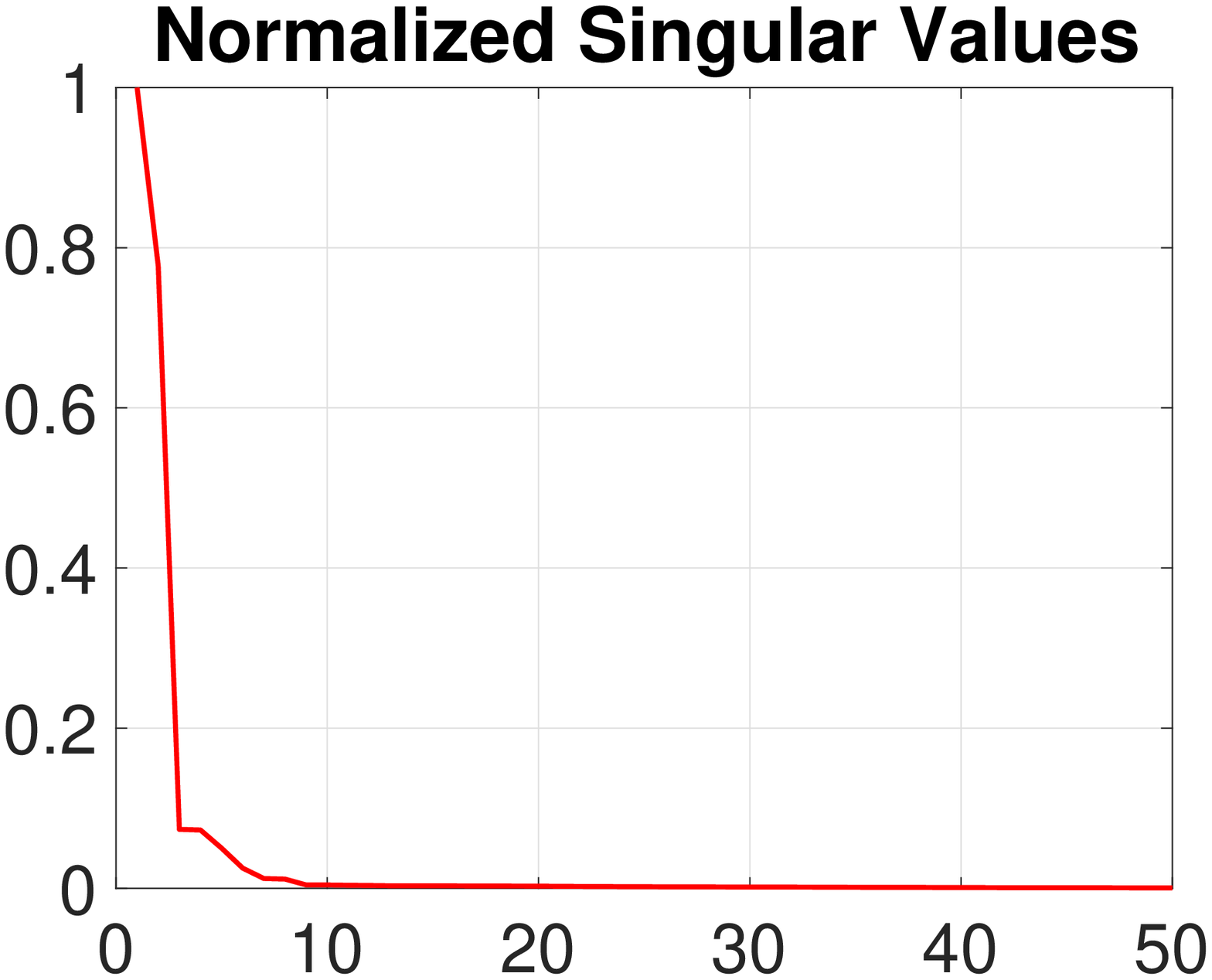}
\includegraphics[width=1.34in]{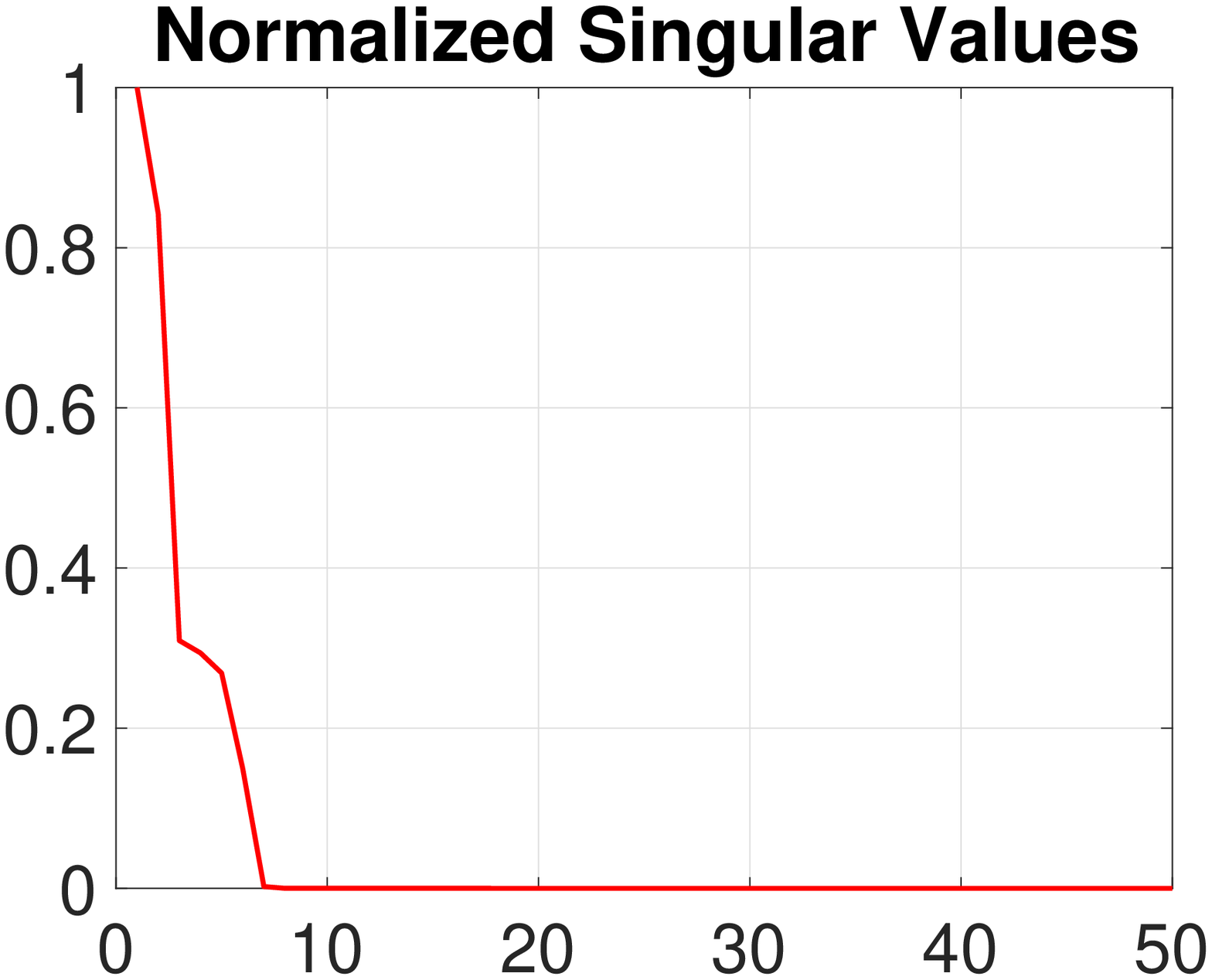}
}
\end{center}
%\vspace{-0.25in}
           \caption{This plot shows 5 normalized time-series where the first  three time-series (from left) are sampled from real data (Yahoo dataset) and the last two time-series are synthetic. The second row demonstrates the normalized eigenvalues of the covariance matrix of the corresponding  trajectory matrices. One can observe that the eigenvalues drop rapidly which suggests that the trajectory matrices can be accurately approximated using a low rank matrix. }    \label{fig:ranks}
           %\vspace{-0.18in}
\end{figure*}

\section{Related Work}
The simplest method to identify outlying time-stamps in a time-series is to treat the  time-stamp values as independent samples from a probability distribution (e.g., Gaussian) and infer the distribution from the former samples. In this paper, we refer to this approach as IID. IID can successfully identify the anomalies whose time-stamp values are outside of the range of the values of normal time-stamps. However, IID is completely blind in identifying the contextual anomalies which do not necessarily push the value outside of the normal rage of the values. It is not a proper assumption in many time-series to presume that time-stamp values are independent as time-series frequently have an underlying structure, for example seasonality. 

A popular approach to utilize the structure of the time-series is to have a running window which provides the opportunity to  compare the local structure of the time-series against the past observations to decide if there is an anomaly in the given window. Some of the window-based methods are Random Cut Forest (RCF) \cite{guha2016robust}, Auto-Regressive (AR) model \cite{braei2020anomaly}, and the Spectral Residual (SR) based method \cite{ren2019time}. 

Another approach to infer the underlying structure of the data is to employ a deep neural network \cite{shen2020timeseries,zhao2020multivariate, su2019robust,hundman2018detecting, munir2018deepant, zhou2019beatgan,xu2018unsupervised, challu2022deep}. However, deep learning methods are not applicable in many applications where there is not enough data to train the network, this is particularly common for low-frequency time-series data. 

\textit{In this paper, we focus on the scenarios where only a short history for a single time-series is available.}

The lines of work most closely related to this paper are  singular spectrum based time-series analysis \cite{dokumentov2014low,golyandina2001analysis,papailias2017exssa,khan2017forecasting}  and  robust matrix decomposition \cite{candes2011robust,chandrasekaran2011rank} which we discuss in the following sections.

\begin{figure*}%[h!]
\begin{center}
\mbox{
\includegraphics[width=1.32in]{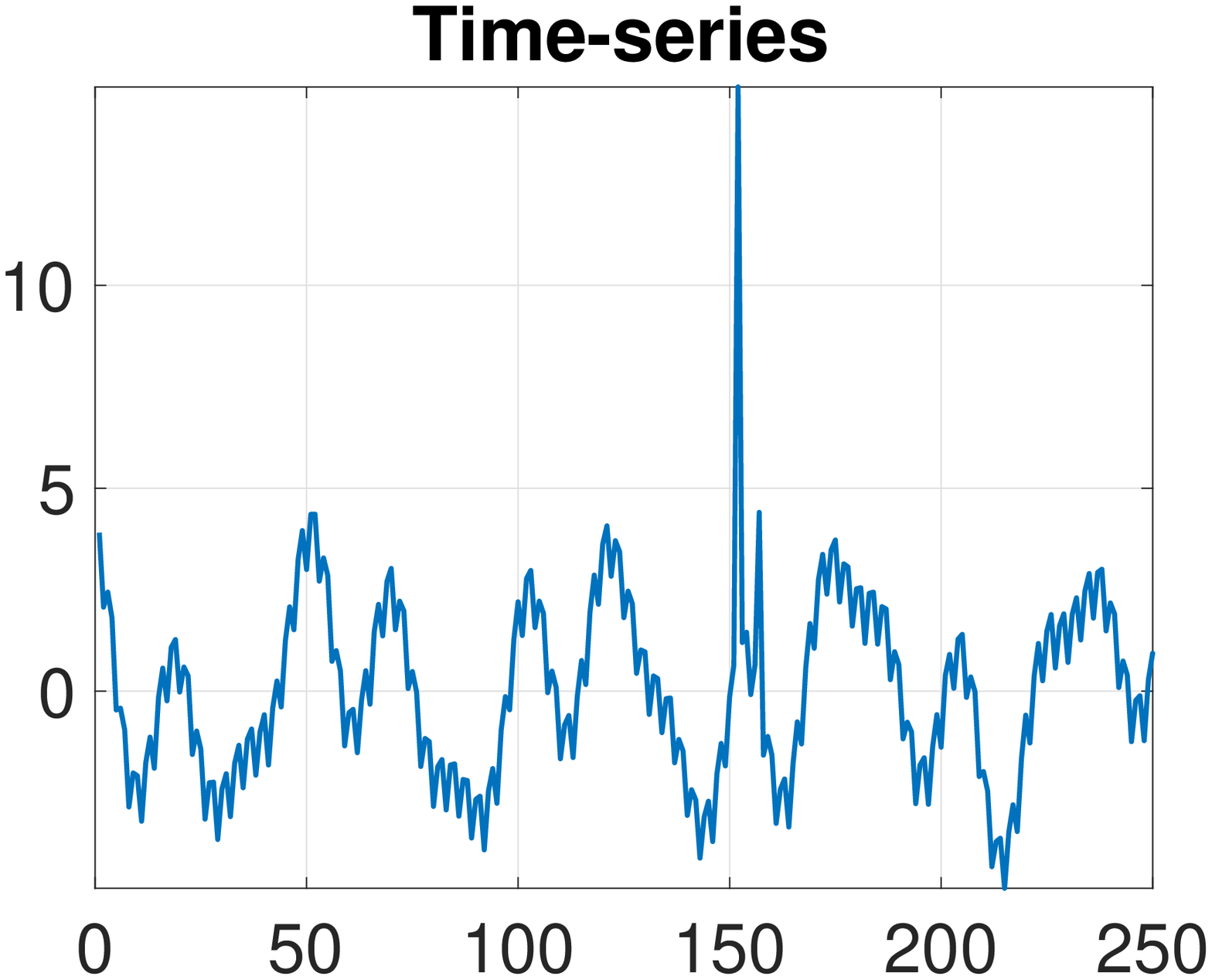}
\includegraphics[width=1.32in]{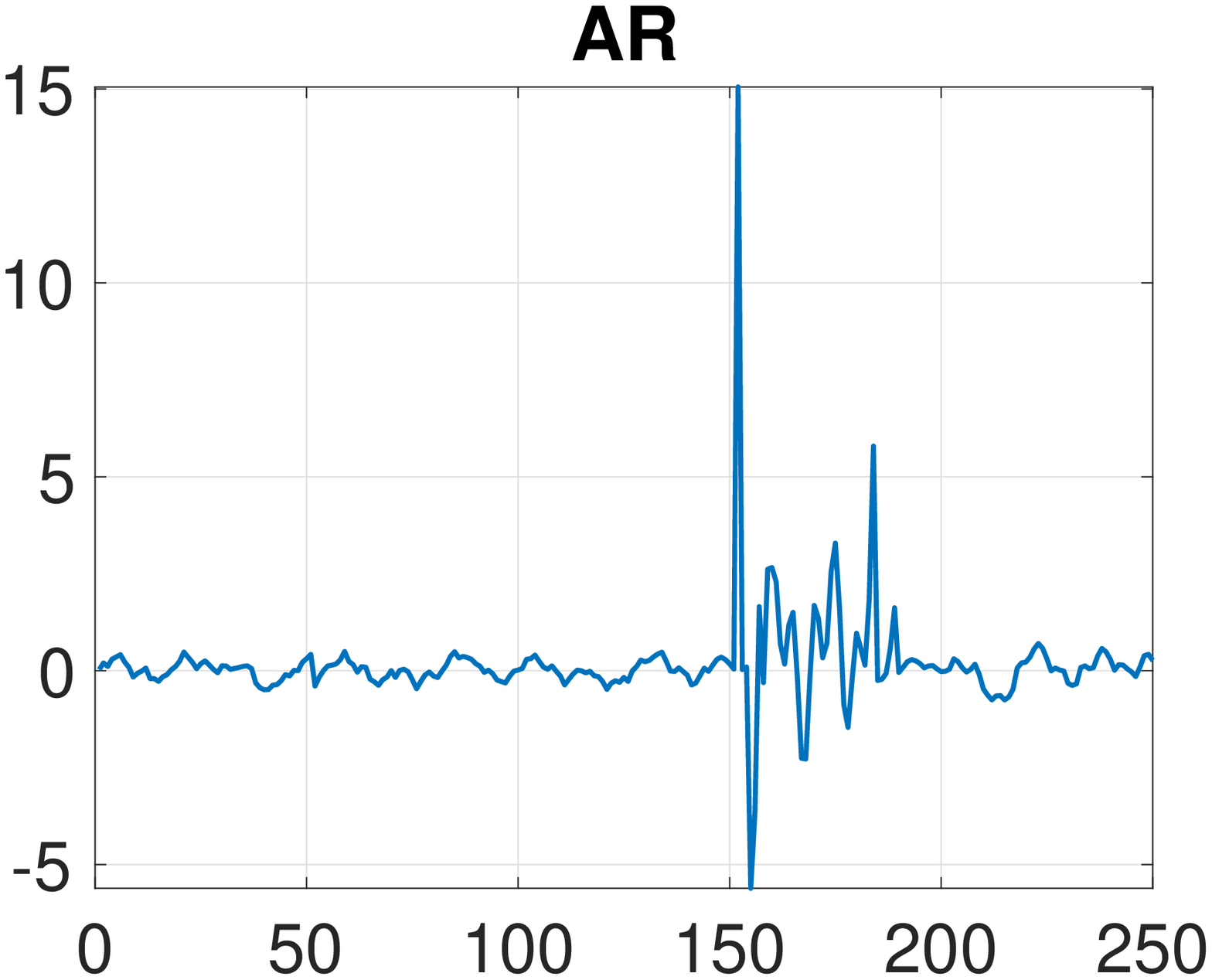}
\includegraphics[width=1.32in]{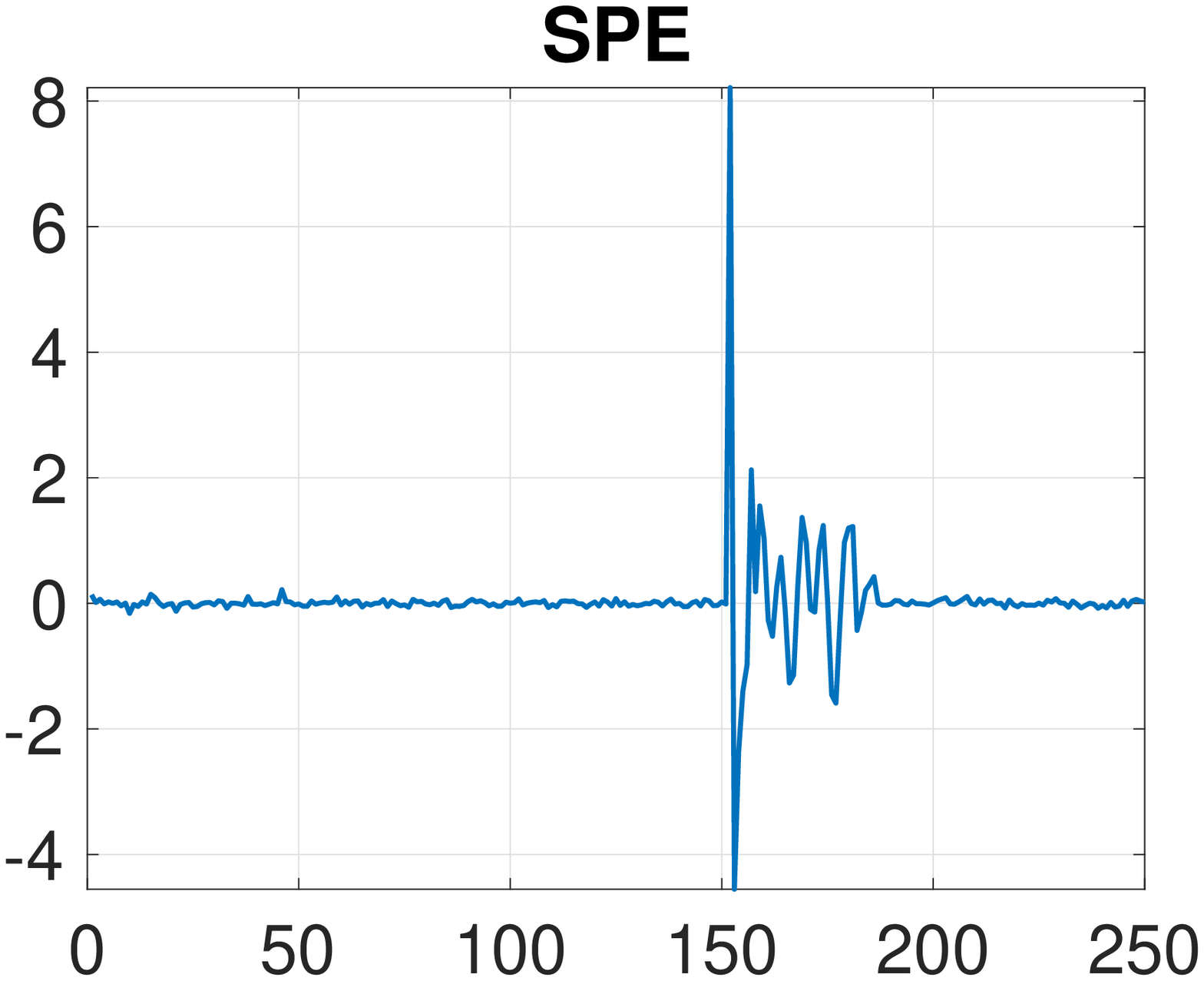}
\includegraphics[width=1.32in]{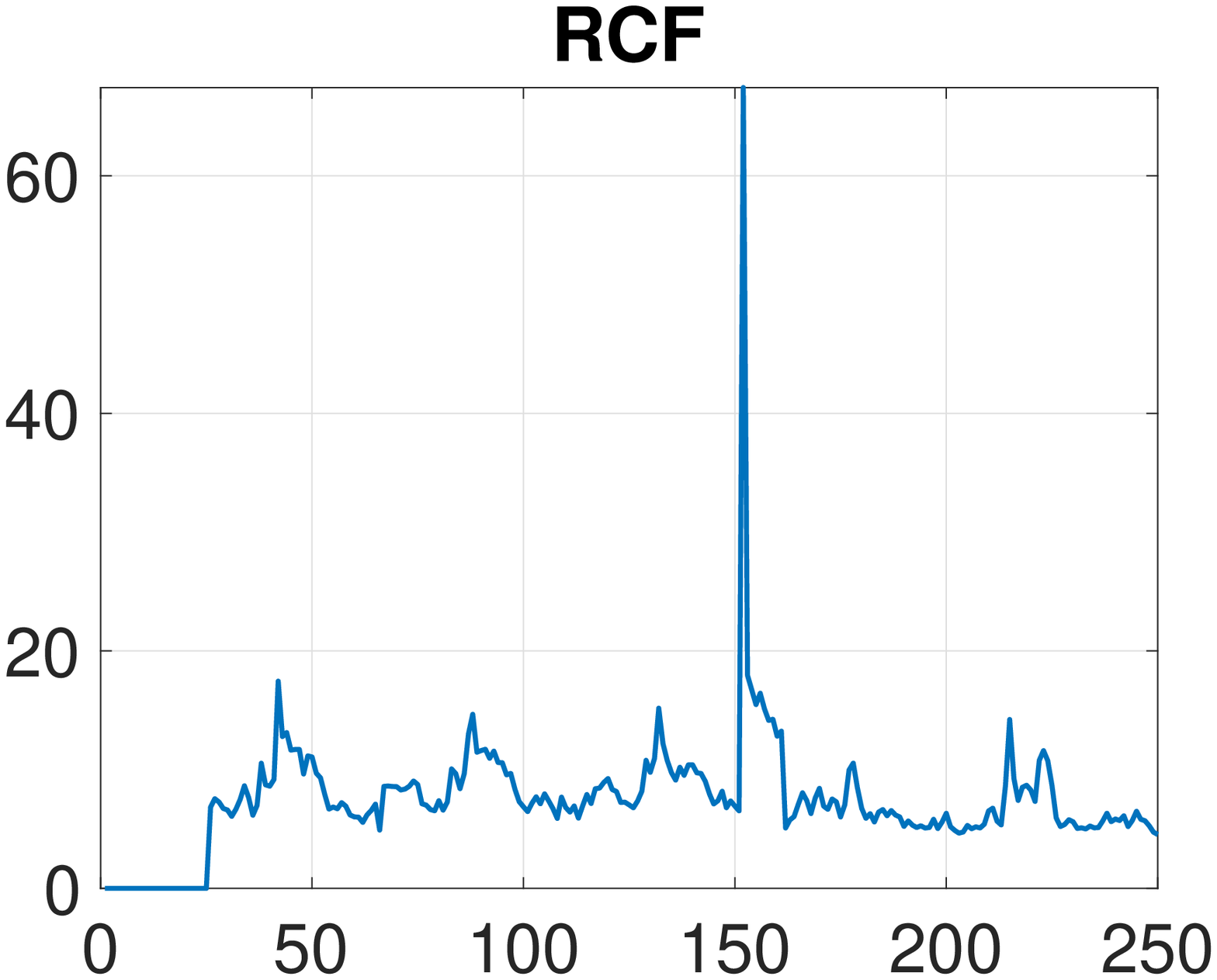}
\includegraphics[width=1.32in]{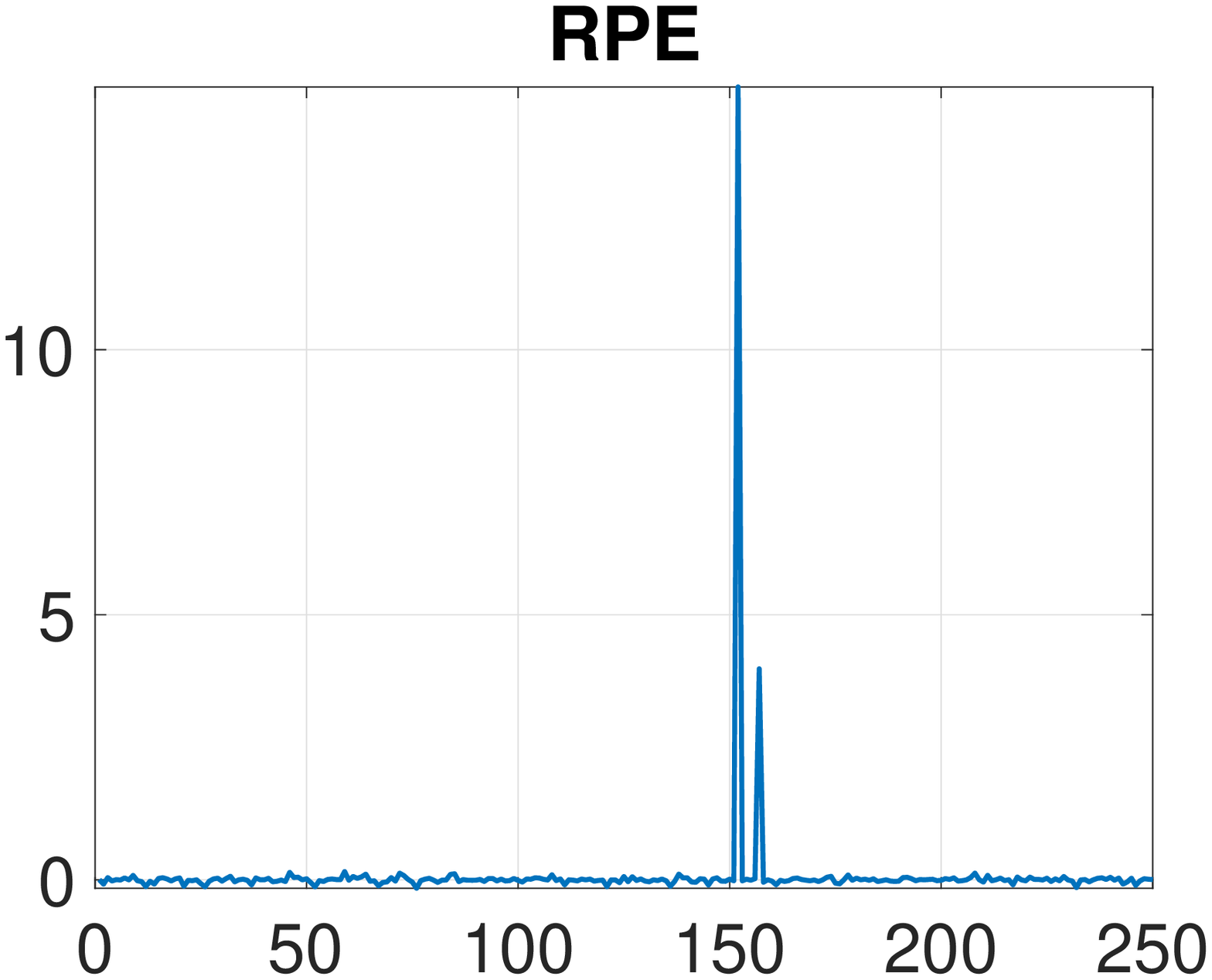}
}
\end{center}
%\vspace{-0.25in}
           \caption{The first plot (from left) shows the time-series which includes two anomalies at indices 151 and 156. The next 4 plots demonstrate the anomaly scores computed by 4 window-based methods. SPE (Simple Projection based anomaly Extraction) is similar to RPE but it employs a simple projection instead of the robust projection step employed in RPE. }    
           \label{fig:resiexample}%\vspace{-0.18in}
\end{figure*}

\subsection{Singular spectrum analysis}
\label{sec:ssa}
Low-rank representations and approximations have been shown to be a very useful tool in
time series analysis. One of the popular approaches is singular spectrum analysis
  that represents the time-series using the trajectory matrix and uses a low-rank approximation
 to model the time-series and the model is used to compute the next values of a time series \cite{dokumentov2014low,golyandina2001analysis,papailias2017exssa,khan2017forecasting}. Singular spectrum analysis uses the fact that many time
series can be well approximated by a class of so-called time series of finite rank \cite{usevich2010signal,gillard2018structured,golyandina2001analysis}. 
Finite rank time-series is a time-series whose trajectory matrix is a low rank matrix. The  real valued time-series of finite rank can be written as the  sums of products of cosines,
exponential and polynomial functions as
$
\bt(k) = \sum_{j=1}^{s_1} Q_j(k) \rho_j^k \cos (2\pi \omega_j k + \phi_j) + \sum_{j=s_1 + 1}^{s_2} Q_j(k) \rho_j^k \:,
$
where $Q_j(k)$ are real polynomials of degrees  $\mu_j - 1$, $\omega_j \in (0, 0.5)$, and $\phi_j \in \mathbb{R}$, $\rho \in \mathbb{R}$ are distinct numbers and  the pairs $\{ (\rho_j, \omega_j) \}_{j=1}^{s_1}$ are also distinct \cite{gillard2018structured}. This class of time-series can accurately model trends, periodicities and
modulated periodicities in time series \cite{golyandina2001analysis}  and the rank of the trajectory matrix is equal to $r = \sum_{j=1}^{s_1} 2\mu_j+ \sum_{j=s_1 + 1}^{s_2} \mu_j$ if $M_1, M_2 > r$ \cite{usevich2010signal}. 
Figure \ref{fig:ranks} shows few real and synthetic time-series along with the singular spectrum of their corresponding trajectory matrices. The curve of the singular values show that the trajectory matrices can be accurately approximated using low rank  matrices. 
The authors of  \cite{gillard2018structured,usevich2016hankel} leveraged the low rank structure of the trajectory matrix to employ matrix completion techniques \cite{candes2009exact,lois2015online} to perform time-series forecasting and missing value imputation. 
%RPE  is related to the Singular spectrum analysis based forecasting methods because it also utilizes the linear structure of the trajectory matrix. However, RPE leverages the linear structure differently as it employs a robust projection algorithm to compute the residual values and the linear structured is learned using a robust subspace recovery algorithm. 

\subsection{Anomaly detection using robust PCA}
An interesting topic in the literature of robust PCA is the problem of low rank plus sparse matrix decomposition where the given data matrix $\bD$ is assumed to be a summation of an unknown low rank matrix and an unknown sparse matrix \cite{candes2011robust,chandrasekaran2011rank,feng2013online}. A convex optimization based solution to this problem was proposed in \cite{candes2011robust,chandrasekaran2011rank} and the authors established sufficient conditions under which the algorithm is guaranteed to decompose the data exactly with high probability. Since the trajectory matrix of time-series with a background signal  can mostly be approximated using a low rank matrix,  one can consider the trajectory matrix of the time-series in the presence of anomalies as the summation of a low rank matrix and a sparse matrix. This observation motivated \cite{jin2017anomaly,wang2018improved} to employ the matrix decomposition algorithm 
analyzed in \cite{candes2011robust,chandrasekaran2011rank,rahmani2017high} for anomaly detection in time-series data. However, the methods proposed in  \cite{jin2017anomaly,wang2018improved} require solving the nuclear norm based optimization problem in each time-stamp whose solver requires computing a SVD of an $M_1 \times M_2$ dimensional matrix per iteration which means that the computation complexity of processing a single time-stamp is $\calO(M_1^2 M_2 k)$ where $k$ is the number of iterations it takes for the solver to converge. In this paper, we also utilize the underlying linear structure of the trajectory matrix but we transform the decomposition problem into a robust projection problem and a closed-form algorithm is presented which saves the algorithm from the complexity of running an iterative solver for every time-stamp. %The computation complexity of the proposed method for processing a  time-stamp is $\calO(M_1 r)$ where $r$ is the 

\section{Motivation: the problem of window-based methods}
Many of the time-series anomaly detection algorithms are window-based methods (e.g., SR, AR, RCF, ...). Although with a window of samples we move the problem into a higher dimensional space where the anomalous pattern are more separable, window-based methods suffer from two main setbacks: \\
$\bullet$ The window-based methods which rely on using an unsupervised outlier detection method (e.g., RCF \cite{guha2016robust}) are partially or completely blind about the location of the anomaly in their window. This shortcoming might cause the algorithm to report a single anomaly for a long period time (as long as the anomaly is present in the window).  
\\
$\bullet$ Some of the window-based methods (such as AR  and SR) uses the samples in the window to compute a residual value corresponding to each time-stamp which is the difference between the actual value and the forecast value. These algorithms are not robust to the presence of anomalies in the window and the presence of anomalies in the window can cause significant error in the computed residual value. %For instance, the residual value computed by AR is equal to $v - \bw^T \bx$ where $v$ is the last time-stamp value, $\bw$ is the learned weight vector, and $\bx$ is the window of past observations. If $\bx$ contains notable anomalies, the computed residual value can be significantly deviated. 

The main contribution of this work is  proposing an efficient and closed-form window-based method which is  robust to the presence of anomalies in its window. Figure \ref{fig:resiexample} shows the anomaly scores computed using 4 window-based methods. The time-series includes two anomalies at indices 151 and 156 and the size of window is equal to 30. One can observe RPE precisely distinguished   the two anomalies and the presence of anomalies did not affect the residual values of the neighboring time-stamps. In contrast, 
with the other methods, we observe two major problems. First, the strong anomaly masked the weaker anomaly; second, the presence of anomalies in the window caused the algorithms to generate high anomaly scores for a long period of time.

\begin{algorithm}
\caption{
RPE for  anomaly detection in univariate time-series}
\label{alg:RPE}

\textbf{Input:} Training time-series $\bt_{tr} \in \mathbb{R}^{n}$, retraining frequency $q$,  window size $M_1$,  and max training size $t_{\max}$.

\smallbreak

\textbf{1. Training} \\
\textbf{1.1} If the length of $\bt_{tr}$ is larger than $t_{\max}$, discard the first $n - t_{\max}$ samples.  \\
\textbf{1.2} % (n-L+1)
Build trajectory matrix $\bX \in \mathbb{R}^{M_1 \times M_2}$ using $\bt_{tr}$.% where $M_2 = n-M_1+1$. 
\\
\textbf{1.3} Apply  the robust subspace recovery algorithm  to $\bX$ to estimate $\calU$ and define $\bU \in \mathbb{R}^{M_1 \times r}$ as an orthonormal basis for $\calU$.\\
\textbf{Remark}. The rank of $\bX$ is estimated by the subspace recovery algorithm.
%\\
%\textbf{1.4} Set $\calM$ as an empty memory and populate it with the residual values of  the samples in $\bt_{tr}$ using the same procedure used in Step 2.2 to Step 2.4.   

\textbf{1.5} Set counter $c = 0$. 

\smallbreak

\textbf{2. Online Inference.} For any new time-stamp value $v$:
\\
\textbf{2.1} Append $v$ to $\bt_{tr}$ and update $c = c + 1$. 
\\
\textbf{2.2} Build $\bx \in \mathbb{R}^{M_1}$ using the last $M_1$ elements of $\bt_{tr}$. 

%using $v$ and the last $L - 1$ received values. 

\textbf{2.2} Use Algorithm 2 (robust projection) to compute $\hat{\ba}$.

\textbf{2.3} Compute the residual value corresponding to $v$ as $e = v - \hat{\ba}^T \bu_{-1}$ where $\bu_{-1}$ is the vector of the last row of $\bU$.

% \textbf{2.4} Append $e$ to memory $\calM$ and define $\vartheta$
% as the empirical CDF value corresponding to $\vartheta$.

%\textbf{2.4} If $\vartheta > \calT$ :\\
%\textbf{2.4.1} Compute $\hat{v} = \hat{\ba}^T \bu_{-1}$.
%\\
%\textbf{2.4.2} Replace the last element of $\bt_{tr}$ with $\hat{v}$.
%\\

\smallbreak
\textbf{2.4} If the reminder of $\frac{c}{q}$ is equal to 0, perform Step 1.1, Step 1.2, and Step 1.3 to update $\bU$.
\end{algorithm}

%\vspace{-0.1in}
\section{Proposed Approach}
%\vspace{-0.1in}
\label{sec:mainprop}
Detecting anomalies in time-series with a background pattern can be challenging since the value of an anomalous  time-stamp can lie in the  range of normal values, which means that thresholding-based methods such as the IID method or the method discussed in  \cite{siffer2017anomaly} are not able to identify the outlying time-stamps. 
In Section \ref{sec:ssa} we discussed how 
 the trajectory matrix of a time-series with a background pattern can mostly be  approximated using a low dimensional linear model \cite{gillard2018structured,papailias2017exssa}. 
The main idea behind the RPE algorithm is to utilize the linear model to extract the outlying component of each time-stamp value. We propose to employ a robust projection step    which makes the presented method robust against the presence of anomalies in its window and it enables the algorithm to distinguish anomalies at the time-stamp level. A closed-form and provable algorithm is presented to perform the robust projection step. 

\begin{remark}
We assume that the given time-series $\bt$ can be written as $\bt = \bt_b + \bt_s + \bt_n$ where $\bt_b$ is the background signal, $\bt_s$ represents the outlying components of the time-stamps ($\bt_s$ is assumed to be a sparse vector) and $\bt_n$ is the additive noise. 
In the next section, it is assumed that the orthonormal matrix  $\bU \in \mathbb{R}^{M_1 \times r}$  defined as the span of the column space of the trajectory matrix of $\bt_b$ is given. Subsequently, we  discuss  efficient algorithms which  can accurately estimate $\bU$ from the trajectory matrix of the training time-series.
\end{remark}

\begin{algorithm}
\caption{Robust Linear Projection}
\label{alg:proj}
%\footnotesize
\textbf{Input:} The orthonormal matrix $\bU \in \mathbb{R}^{M_1 \times r}$, the observation vector $\bx$, and $n_s$ as an upper-bound on the number of corrupted elements of $\bx$.

\textbf{1.} Define $\hat{\be} = | \bx - \bU \bU^T \bx|$.

\textbf{2.} Define set $\calI$ with cardinality $M_1 - n_s$ as the indices of the elements of $\hat{\be}$ with the smaller values. 

\textbf{3.} Define $\bU_{\calI} \in \mathbb{R}^{(M_1 - n_s) \times r}$ as the rows of $\bU$ whose indices are in $\calI$ and define $\bx_{\calI}$ similarly. 

\textbf{Output:}  $\hat{\ba} = (\bU_{\calI}^T \bU_{\calI})^{-1} \bU_{\calI}^T \bx_{\calI} $.
\end{algorithm}

%\vspace{-0.1in}
\subsection{Robust projection for anomaly detection in time-series}
%\vspace{-0.1in}
\label{sec:originalidea}
It is assumed that the trajectory matrix of $\bt_b$ can be accurately described using a low dimensional linear subspace $\calU$ spanned by $\bU$. Therefore, each window $\bx \in \mathbb{R}^{M_1}$ can be written as 
 \begin{eqnarray}
  \begin{aligned}
\bx = \bU \ba + \bs + \bn \:,
  \end{aligned}
\end{eqnarray}
where $\bs$ and $\bn$ represent the components corresponding to anomalies and noise, receptively. In order to simplify the following discussion, let us temporarily ignore the presence of the added noise and presume that
 \begin{eqnarray}
  \begin{aligned}
\bx = \bU \ba + \bs \:.
  \end{aligned}
\end{eqnarray}
We would like to estimate $\bs$ to identify the anomalous time-stamps.  A simple solution is to simply project $\bx$ into the column space of $\bU$ which means that $\bs$ is estimated as $\hat{\bs} = \bx - \bU \hat{\ba}$ where $\hat{\ba}$ is the optimal points of 
 \begin{eqnarray}
  \begin{aligned}
\arg\min_{\dot{\ba}} \| \bx - \bU \dot{\ba} \|_2 = \bU^T \bx \:.
  \end{aligned}
  \label{eq:svd}
\end{eqnarray}
The algorithm based on (\ref{eq:svd}), which we call it Simple Projection based anomaly Extraction (SPE), is still one of the best algorithms for anomaly detection in the time-series with a back-ground signal. However,
 the estimation obtained by  (\ref{eq:svd}) is the optimal point of the likelihood function when we model the elements of $\bs$ as independent samples from a zero mean Gaussian distribution.  This is an obviously wrong assumption given that $\bs$ is known to be a sparse vector (i.e., the anomalies corrupt few time-stamps) and the non-zero elements of $\bs$ can have  arbitrarily large values. It is well known that $\ell_1$-norm is  robust to the presence of sparse corruptions \cite{candes2005decoding,candes2011robust}. The authors of \cite{candes2005decoding}
provided the sufficient conditions which guarantee that the optimal point of 
 \begin{eqnarray}
  \begin{aligned}
\arg\min_{\dot{\ba}} \| \bx - \bU \dot{\ba} \|_1
  \end{aligned}
  \label{eq:elaa}
\end{eqnarray}
is equal to $\ba$. The key factors to ensure the success of (\ref{eq:elaa}) are that the column-space of $\bU$ should not include sparse vectors and $\bs$ should be sufficiently sparse. In order to show the dependency  on these two important factors, we derived the following  new guarantee whose sufficient condition is more insightful  in showing the dependency on the incoherency of $\calU$ (the proof is also  easier). In the next section, we show that the incoherency of $\calU$ can be leveraged to design a provable closed-form algorithm. 

\begin{lemma}
Suppose vector $\bx \in \mathbb{R}^{M_1}$ is generated as $\bx = \bU \ba + \bs$ where $\bU \in \mathbb{R}^{M_1 \times r}$ and vector $\bs$ contains $m$ non-zero elements. If  \begin{eqnarray}
  \begin{aligned}
\kappa (\bU)< \frac{1}{2 m \: \sqrt{r}}\:,
  \end{aligned}
  \label{eq:58}
\end{eqnarray}
then $\ba$ is the optimal point of (\ref{eq:elaa}).
\label{lm:my_candes}
\end{lemma}

\noindent
Lemma \ref{lm:my_candes} indicates that if $\calU$ is sufficiently incoherent and $\bs$ is sparse enough, then $\ba$ can be perfectly recovered, no matter how large the non-zero elements of $\bs$ are. This feature enables the anomaly detection algorithm to achieve high accuracy in the point-wise detection of anomalous time-stamps. In online anomaly detection, the last element of $\bx - \bU \hat{\ba}$ is the residual value of the last received time-stamp. Lemma \ref{lm:my_candes} indicates that \textit{even if the window contains $m-1$ anomalies, the residual value computed by (\ref{eq:elaa}) is not influenced by them}. This is the main motivation that in this paper we propose to employ a robust projection algorithm to compute the residual values. 

%\vspace{-0.1in}
\subsection{The closed-form solver for the robust projection step}
%\vspace{-0.1in}
In section \ref{sec:originalidea}, we proposed to utilize the linear structure of the trajectory matrix and employ the robust projection based on (\ref{eq:elaa}) to compute the residual values since it ensures that an anomaly only affects the residual value of its corresponding time-stamp which leads to achieving high detection accuracy. However, the optimization problem (\ref{eq:elaa}) does not have a closed-form solution and one needs to employ an iterative solver to find the optimal point for each time-stamp. %A non-iterative algorithm with closed form steps is preferred specially in the production phase where we require fast inference. 

The following lemma reveals the key idea to design a  closed form solution. It shows that if $\calU$ is sufficiently incoherent and $\bs$ is sparse enough, then the residual values computed based on (\ref{eq:svd}) can be used to find a set of candidate indices of the corrupted time-stamps. 

\begin{remark}
In the following lemma, in order to simplify the exposition of the result, it is assumed that the absolute value of the non-zero elements of $\bs$ are equal. This assumption is not important in the performance of the algorithm and the full result without the assumption along with the proof is available in the appendix. 
\end{remark}

\begin{lemma}
Suppose $\bx$ is generated as in Lemma \ref{lm:my_candes}, assume the absolute value of the non-zero elements of $\bs$ are equal to $\tau$,  define $\hat{\be} = \left| \bx - \bU \bU^T \bx \right|$, and define $\calI_s$ as the set of indices of the non-zero elements of $\bs$. If 
  \begin{eqnarray}
  \begin{aligned}
\mu^2(\bU) \le \frac{1}{2 \: r\: m}  \:,
  \end{aligned}
  \label{eqLsuff}
\end{eqnarray}
then 
$
\min_{i \in \calI_s} \hat{\be}(i)  > \max_{i \notin \calI_s} \hat{\be} (i)\:,
$ where $m=|\calI_s|$.
\label{lm:closed-form}
\end{lemma}

\noindent
Importantly, Lemma \ref{lm:closed-form} indicates that although the residual values computed using (\ref{eq:svd}) are faulty  (a single anomaly can cause significant residual values in all the time-stamps of the window), they can be used to identify the potentially corrupted time-stamps.  Based on Lemma \ref{lm:closed-form}, the following corollary can be concluded. 

\begin{corollary}
Define $\hat{\be}$ as in Lemma \ref{lm:closed-form} and define set $\calI_c$ with $|\calI_c| = n_s > m$ as the set of indices of  $\hat{\be}$ with the largest elements where $|\calI_c| $ is the cardinality of $\calI_c$. Form matrix $\bU_{\calI_c^{\perp}} \in \mathbb{R}^{(M_1 - n_c) \times r}$ using the rows of $\bU$ whose indices are not in $\calI_c$ and define $\bx_{\calI_c^{\perp}}$ similarly.
In addition, suppose $\text{rank}(\bU_{\calI}) =  r$ for any set of indices with $|\calI| = n_s$.
 If (\ref{eqLsuff}) holds, then
  \begin{eqnarray}
  \begin{aligned}
({\bU_{\calI_c^{\perp}}}^T \bU_{\calI_c^{\perp}})^{-1} {\bU_{\calI_c^{\perp}}}^T \bx_{\calI_c^{\perp}} = \ba \:.
  \end{aligned}
\end{eqnarray}
\label{col:1}
\end{corollary}

It is important to note that $\tau$ does not appear in the sufficient condition which means no matter how large are the anomalies, the proposed method is guaranteed to recover $\ba$ correctly. Lemma \ref{lm:closed-form} and Corollary \ref{col:1}
showed that the incoherency of $\bU$ imposed an structure on the residual values computed using (\ref{eq:svd}) such that we can filter out the potentially corrupted observations and compute the correct $\ba$ in a closed form way. 
The following theorem states that the estimation error is proportional to the variance of the noise if the observation vector $\bx$ contains additive noise. % noise which means that no matter how large are the anomalies, the estimation error remains bounded where the bound depends on the noise variance. 

\begin{theorem}
Suppose $\bx = \bU \ba + \bs + \bn$ where $\bU$ and $\bs$ are as in Lemma \ref{lm:closed-form} and $\bn \sim	\calN(\textbf{0}, \sigma_n^2 \bI)$ is the added Gaussian noise and define  $\eta$  such that $\mathbb{P}[ | \bn(i) | > \eta] < \frac{\delta}{M_1}$. In addition, suppose $\text{rank}(\bU_{\calI}) =  r$ for any set of indices with $|\calI| = n_s$ where $n_s > m$ and $m = |\calI_s|$.
Define $\bU_{\calI_c^{\perp}}$ as in Corollary \ref{col:1} and define $\hat{\ba} = ({\bU_{\calI_c^{\perp}}}^T \bU_{\calI_c^{\perp}})^{-1} {\bU_{\calI_c^{\perp}}}^T \bx_{\calI_c^{\perp}}$.
 If 
 \begin{eqnarray}
  \begin{aligned}
\mu^2 \left(1 + \frac{\eta}{m \tau} \sqrt{\frac{M_1 \log 1/ \delta}{2}} \right) \le \frac{1}{2 \: r \: m}  \:,
  \end{aligned}   
\end{eqnarray}
then $\hat{\ba}$ is an unbiased estimate of $\ba$ and the variance of $\hat{\ba}(i)$ is smaller than $\sigma_n$ with probability at least $1 - 2\delta$.
\label{thm:withnoise}
\end{theorem}

\noindent
The sufficient condition  states that $\bU$ should be sufficiently incoherent with the standard basis. Matrix $\bU$ spans the column space of the trajectory matrix $\bX$ and it means that RPE can successfully compute the residual values if the given time-series is not sparse (the column space of $\bX$ does not contain sparse vectors). In addition, the sufficient condition requires $m=|\calI_s|$ to be small enough which means that the corrupted elements should not dominate $\bx$.

Algorithm \ref{alg:RPE} presents the RPE algorithm  and Algorithm  \ref{alg:proj} demonstrates the  closed form robust projection step. Algorithm \ref{alg:proj} uses $n_s$ as an upper-bound on the number of anomalies which can be present in window $\bx$. Normally a small portion of the time-stamps are anomalous and a small value for $n_s$ (e.g., $0.05 \: M_1$) is sufficient. %Although with choosing a high value for $n_s$ we can be more confident about filtering out the outlying time-stamps, but a large $n_s$ can cause the rank of $\bU_{\calI_c^{\perp}}$ to be smaller than $r$.

Algorithm \ref{alg:RPE} includes a retraining logic which uses the accumulated samples of the time-series to update the estimation of $\bU$. This is  helpful for the low data regimes to keep improving the estimate of $\bU$ and it can be ignored if the given $\bt_{tr}$ is long enough. 
In our experiments, we turned off the retraining logic when the length of $\bt_{tr}$ becomes larger than $10 M_1$. 
In addition, in the special scenarios where the background pattern could change through time, the retraining logic can be employed to update $\bU$ with a proper frequency. 

Similar to several other window-based methods, the output of Algorithm 1 is the computed residual values. In this paper, we directly apply a threshold to the computed residual values to compute the accuracy of the model. One can convert the computed residual values to probabilistic scores via inferring the distribution of the generated residual values. In addition, in the applications where the user can provide feedback, dynamic thresholding can be utilized to learn the user's desired detection rate \cite{raginsky2012sequential}.

%\vspace{-0.1in}
\subsubsection{Robust estimating of $\calU$:}

In the presumed data model, we assumed that the given time-series can be written as $\bt = \bt_b + \bt_s + \bt_n$ which  means that $\bX = \bB + \bS + \bN$ where $\bX$ is the trajectory matrix of $\bt$ and $\bB$, $\bS$, and $\bN$ are the components corresponding to $\bt_b$, $\bt_s$, and $\bt_n$, respectively. Since $\bB$ is a low rank matrix for most of the  background signals, the robust matrix decomposition algorithm studied in \cite{chandrasekaran2011rank} can be employed to recover $\bB$ from $\bX$ and $\calU$ can be estimated as the column space of the recovered $\bB$. However, the method proposed in \cite{chandrasekaran2011rank} involves solving an optimization problem which requires running an iterative solver for many iterations. In time-series, a small portion (mostly less than 1 $\%$) of the time-stamps are anomalies and in this  section, we discuss techniques which leverage the sparsity of the anomalies to estimate $\calU$ efficiently. We discuss three approaches  to filter out the outlying components from $\bX$ to estimate the column space of $\bB$ from the filtered/cleaned $\bX$. 

\begin{figure}%[h!]
\begin{center}
\mbox{
\includegraphics[width=1.45in]{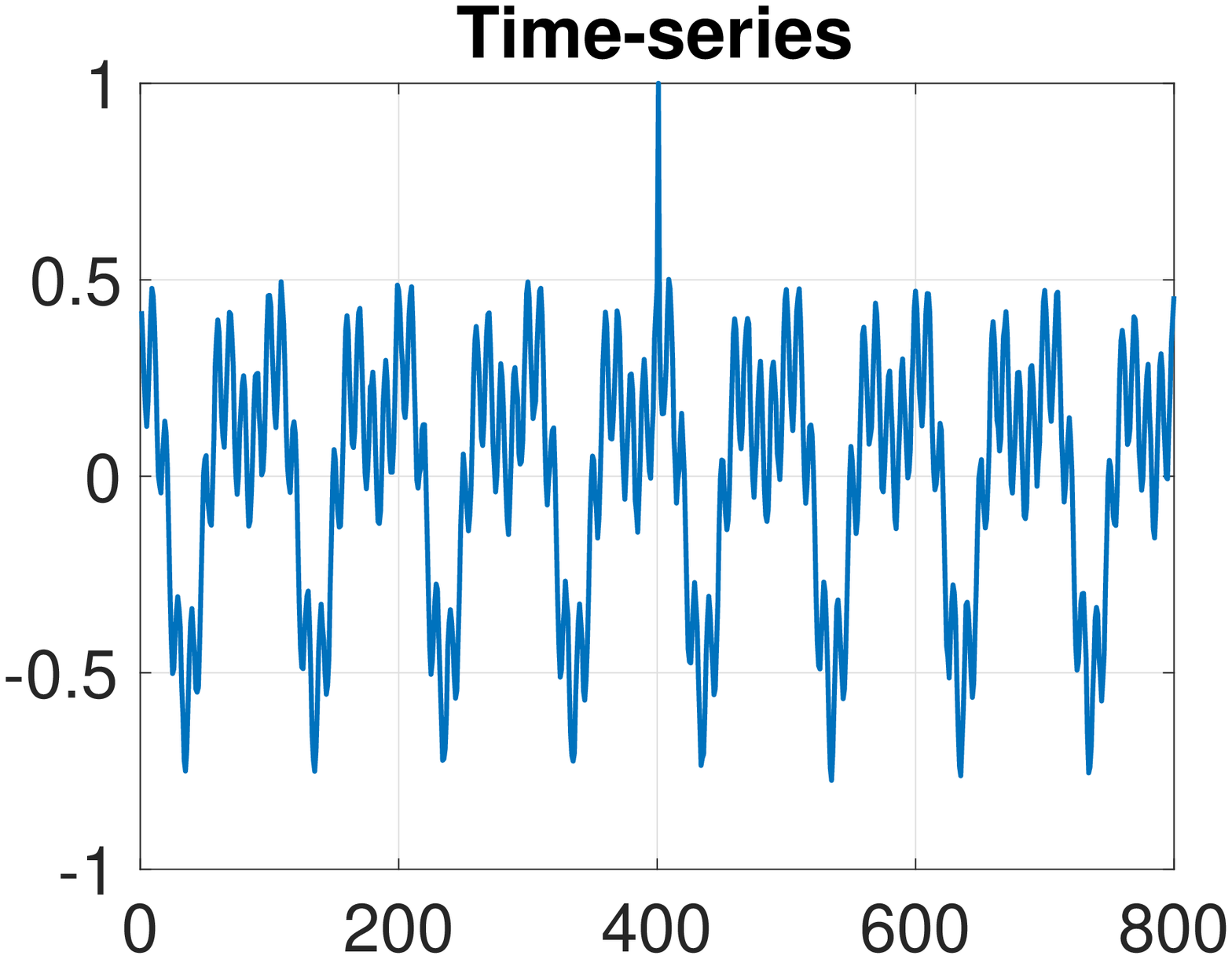}
\includegraphics[width=1.45in]{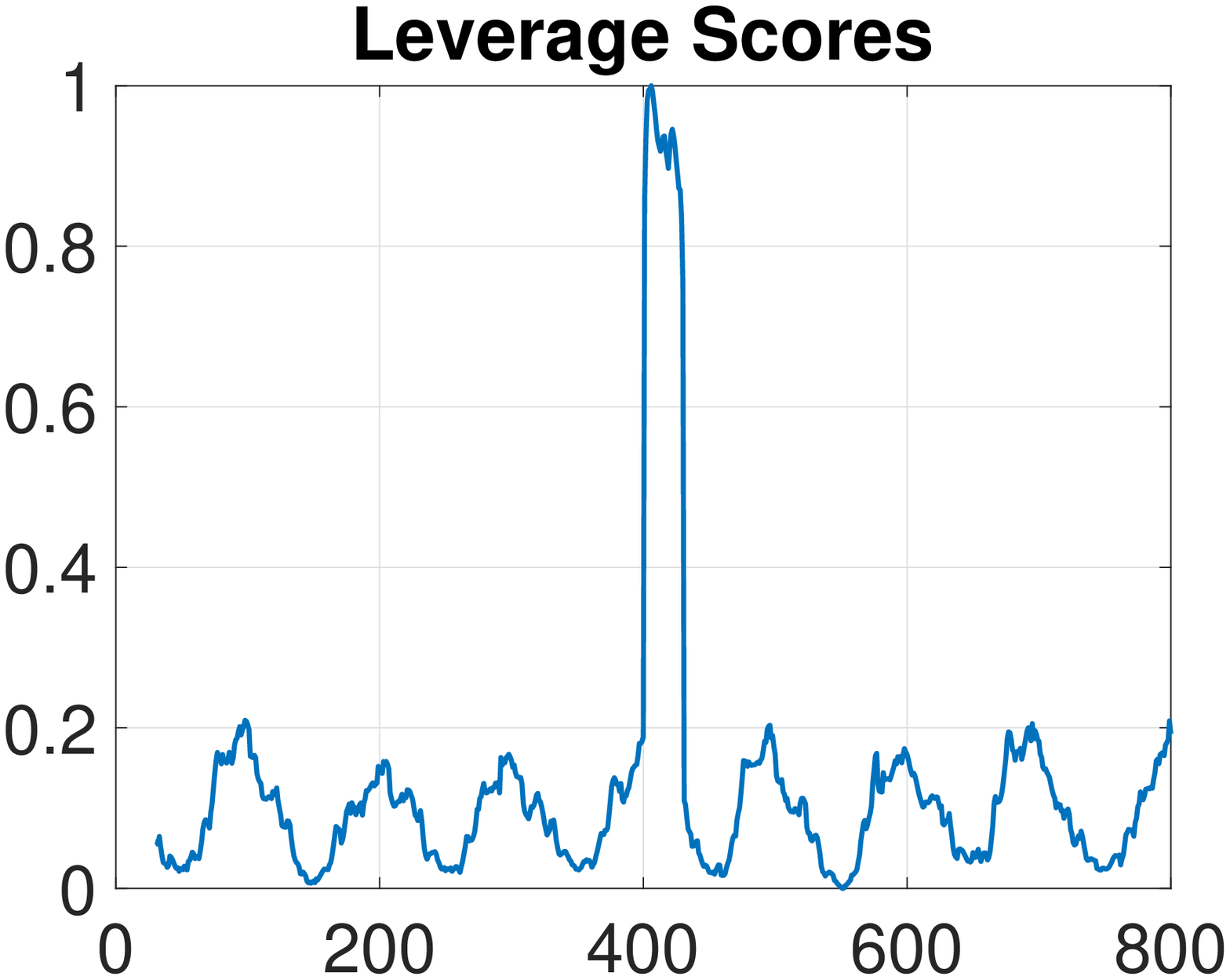}
}
\end{center}
%\vspace{-0.25in}
           \caption{The left plot shows a time-series which includes one  anomaly. The right plot shows leverage scores where each score corresponds to one column of the trajectory matrix.  In this example, if we filter out a small portion of the columns of $\bX$ corresponding to the largest leverage scores, $\bX$ will be free of outlying columns.  }    
           \label{fig:example_leverage}\vspace{-0.18in}
\end{figure}

$\bullet$ \textit{Column-wise approach}: 
In the scenarios where $n \gg M_1$ and a small portion of the elements of $\bt$ are anomalies,  a small portion of the columns of $\bX$ contain the outlying time-stamps and we can employ a column-wise anomaly detection algorithm such as leverage score or the methods proposed in \cite{lerman2018fast,rahmani2019outlier,rahmani2017coherence,rahmani2021fast} to identify them (the columns which contain anomalies) and filter them from $\bX$. The remaining clean columns can be used to estimate $\bU$. If we utilize leverage scores to identify the outlying columns of $\bX$, the computation complexity of this step is $\calO(M_1 M_2 r)$. Figure \ref{fig:example_leverage} shows an illustrative example.

$\bullet$ \textit{Element-wise approach}: 
The first discussed technique was based on the idea of filtering out the outlying columns of the trajectory matrix. However, when $n$ is not significantly larger than $M_1$, a single outlying time-stamp appears in a large portion of the columns of $\bX$ and one can not filter out the anomalies via removing a small portion of the columns of $\bX$. The second technique is based on identifying the outlying elements of $\bX$ and replacing them with the estimated values. 
Algorithm 3 leverages the fact that when a small portion of the elements of $\bX$ are anomalies, the span of the dominant left singular vectors of $\bX$ are close to $\calU$ (the column space of $\bB$). Algorithm 3 replaces a small portion of the elements of $\bX$ (the elements with the largest absolute values) with the median value to make the initial estimate of $\calU$ robust to arbitrary large anomalies. Next, the algorithm utilizes the estimated $\calU$ to identify the candidate outlying elements to update the estimation of $\calU$. The computation complexity of Algorithm 3 is $\calO(M_1 M_2 r)$.

$\bullet$ \textit{Simple approach:} In most of applications, a small portion of the time-stamp values are outliers (mostly less than 1 $\%$). However, the value of the outlying time-stamps could be significantly large such that their presence in $\bX$ could  deviate the span of the major principal components of $\bX$
 from $\calU$. Contextual anomalies have a negligible impact on the major principal components of $\bX$ and in order to tackle the impact of the anomalies with large time-stamp values, Algorithm 4 replaces a small portion of the time-stamp values (which have the largest absolute values) with the median value of the given time-series (median is robust to the presence of anomalies). Although we are not replacing those values with the correct values, its impact on the performance of the algorithm is negligible since we are replacing a small portion of the time-stamp values ($\beta$ can be chosen less than 1 in most applications). This simple technique can ensure that $\bX$ does not contain outlying elements with arbitrary large values and we can rely on the principal components of $\bX$ to obtain $\bU$. 
 In our investigations, we observed that this simple method (Algorithm 4) performs as well as Algorithm 3 in most cases and we used it in the presented experiments to estimate $\calU$. 
 %Although Algorithm 3 handles the presence of anomalies   

\begin{remark}
Different methods can be used to choose $r$. In this paper, we set $r$ equal to number of the singular values which are greater than $s_1/100$ where $s_1$ is the first singular value. Another approach is to choose the rank based on $\frac{\| \bX - \bU \bU^T \bX \|_F}{\| \bX \|_F}$ where $\| \bX \|_F$ is the Frobenius norm of $\bX$.
\end{remark}

\begin{algorithm}
\caption{Robust estimation of $\calU$ (element-wise approach)}
\label{alg:U}
{%\footnotesize
\textbf{Input:} The given time-series $\bt$ and parameter $\alpha$.

\textbf{1.} Form $\bX$ as the trajectory matrix of $\bt$.

\textbf{2.} Select  $\alpha$ percent of the elements of $\bt$ with highest absolute values and replace their values with the median  value  $\bt$. In addition, define $\bQ$ as the trajectory matrix of the updated $\bt$.

\textbf{3.} Define $\bU$ as the matrix of first $r$ left singular vectors of $\bQ$.

\textbf{4} Replace the potential anomalies with the estimated values: \\
\textbf{4.1} Define $\bE = | \bX - \bU\bU^T \bX|$. \\
\textbf{4.2} Define $\calI_o$ as the set of indices of $\alpha$ percent of the elements of $\bE$ with the highest values. 
\\
\textbf{4.3} Replace the elements of $\bX$ whose indices are in $\calI_o$ with the corresponding values in matrix $\bU \bU^T \bQ$.
\\
\textbf{5.} Update $\bU$ as the matrix of first $r$ left singular vectors of  the updated $\bX$.
 
 \textbf{Output:} Matrix $\bU$. 
 }
\end{algorithm}

\begin{algorithm}
\caption{Robust estimation of $\calU$ (simple approach)}
\label{alg:Usimple}
{
%\footnotesize
\textbf{Input:} The given time-series $\bt$ and parameter $\beta$ (e.g., $\beta = 1$).

\textbf{1.} Define $\calI_m$ as the indices of $\beta \: \%$ of elements of $\bt$ with the largest absolute values.

\textbf{2.} Replace the values of the elements of $\bt$ whose indices are in $\calI_m$ with the median value of $\bt$.

\textbf{3.} Form $\bX$ as the trajectory matrix of $\bt$.

\textbf{4.} Define $\bU$ as the matrix of first $r$ left singular vectors of $\bX$.
 
 \textbf{Output:} Matrix $\bU$. 
 }
\end{algorithm}

\begin{figure*}%[h!]
\begin{center}
\mbox{
\includegraphics[width=1.32in]{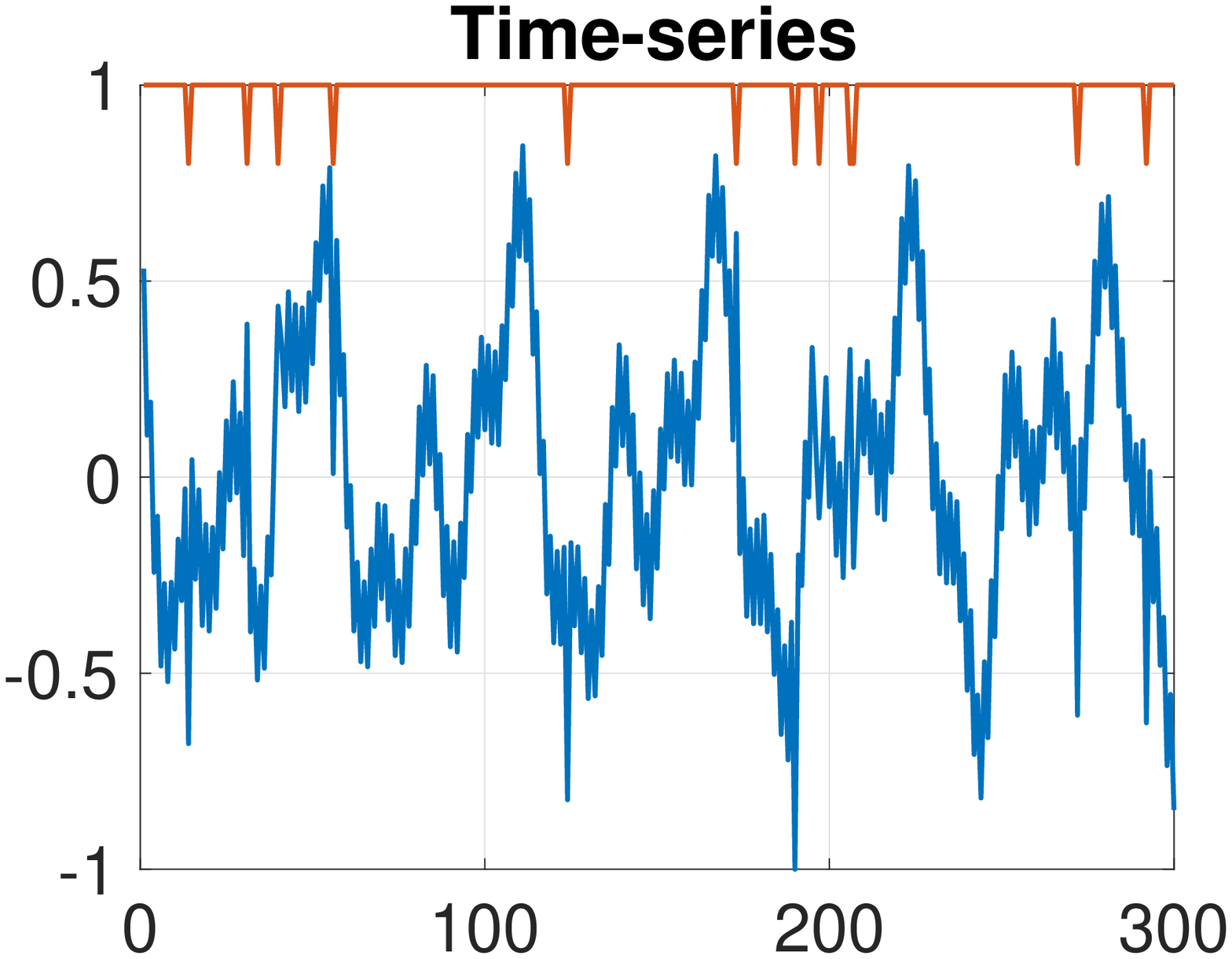}
\includegraphics[width=1.32in]{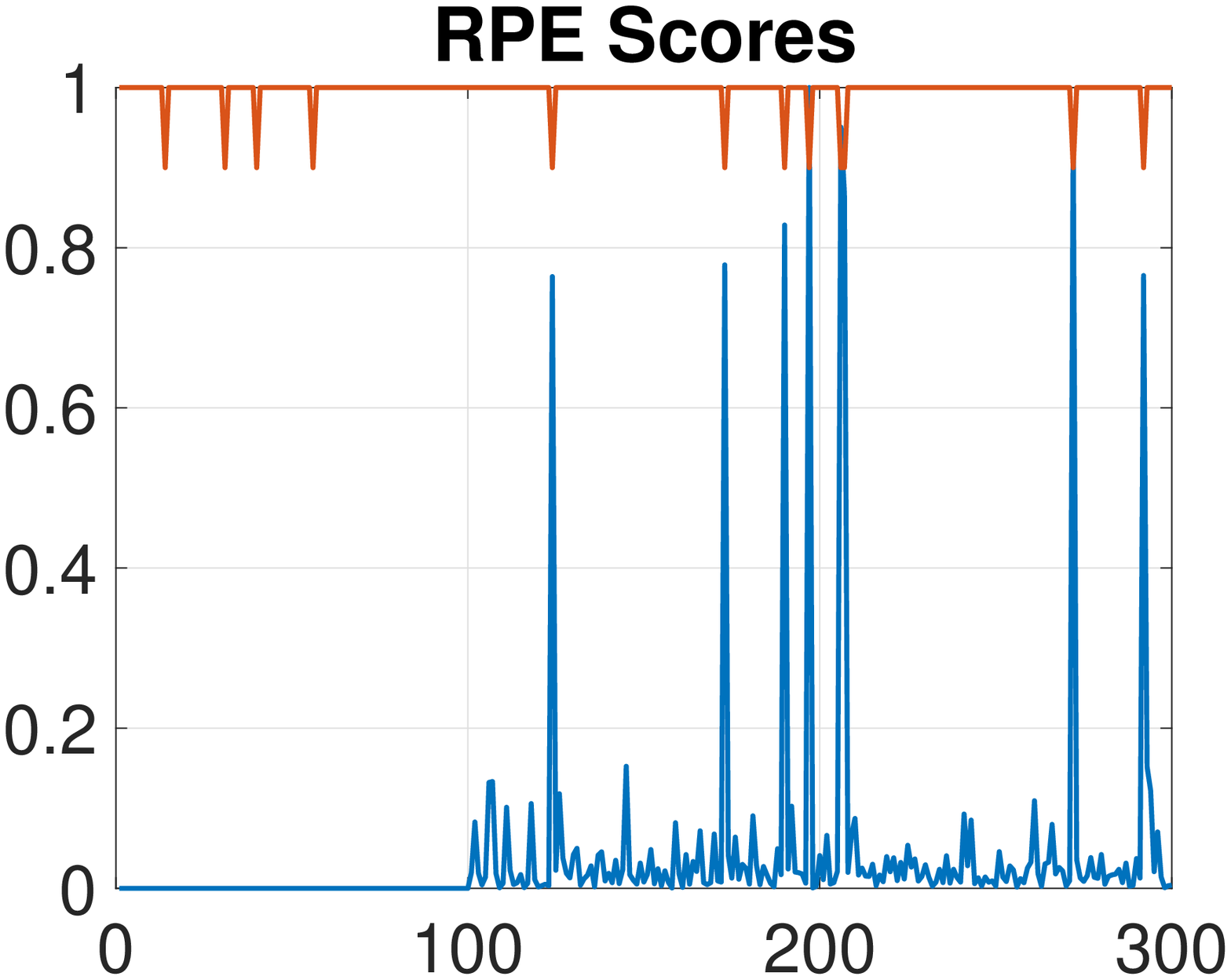}
\includegraphics[width=1.32in]{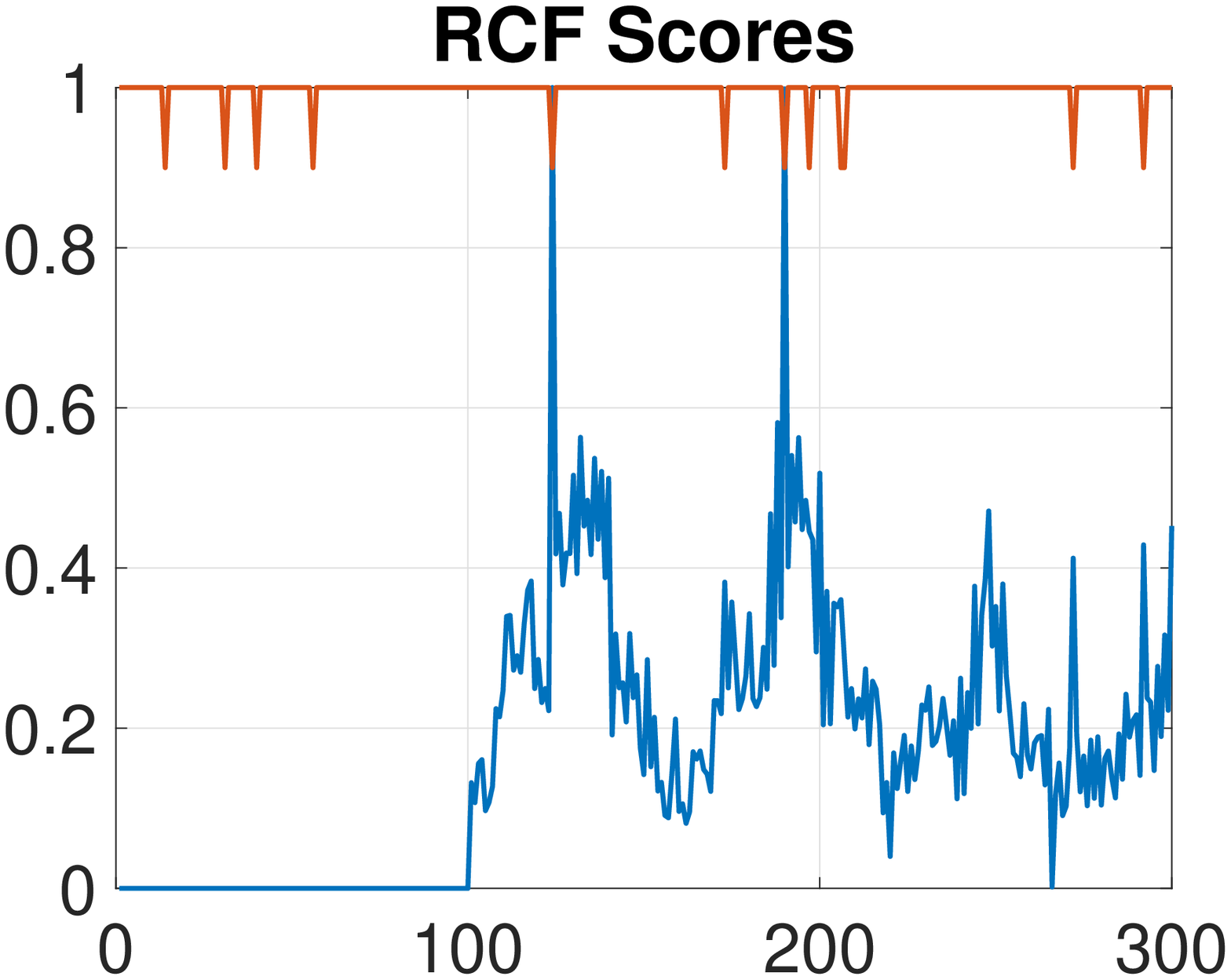}
\includegraphics[width=1.32in]{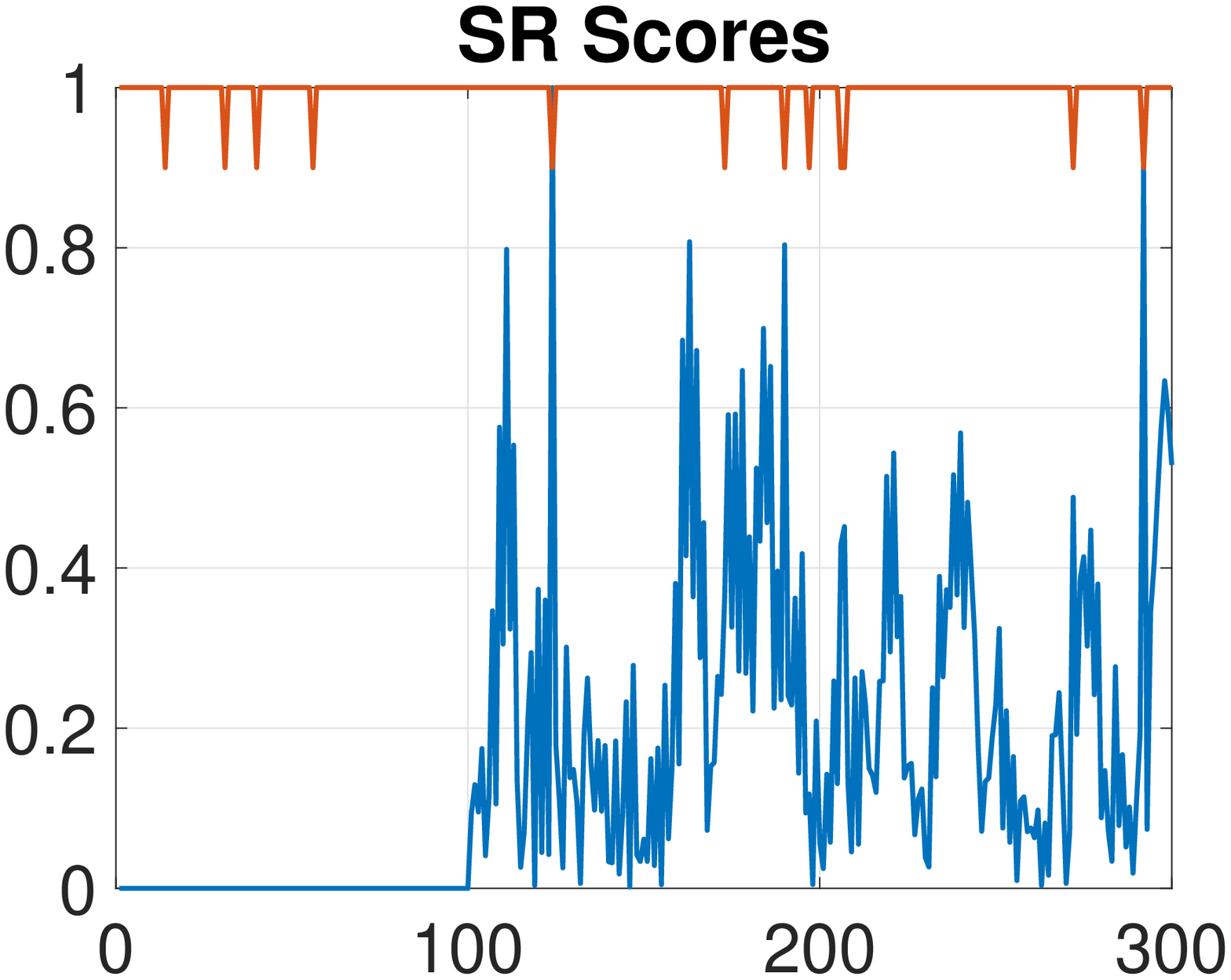}
\includegraphics[width=1.32in]{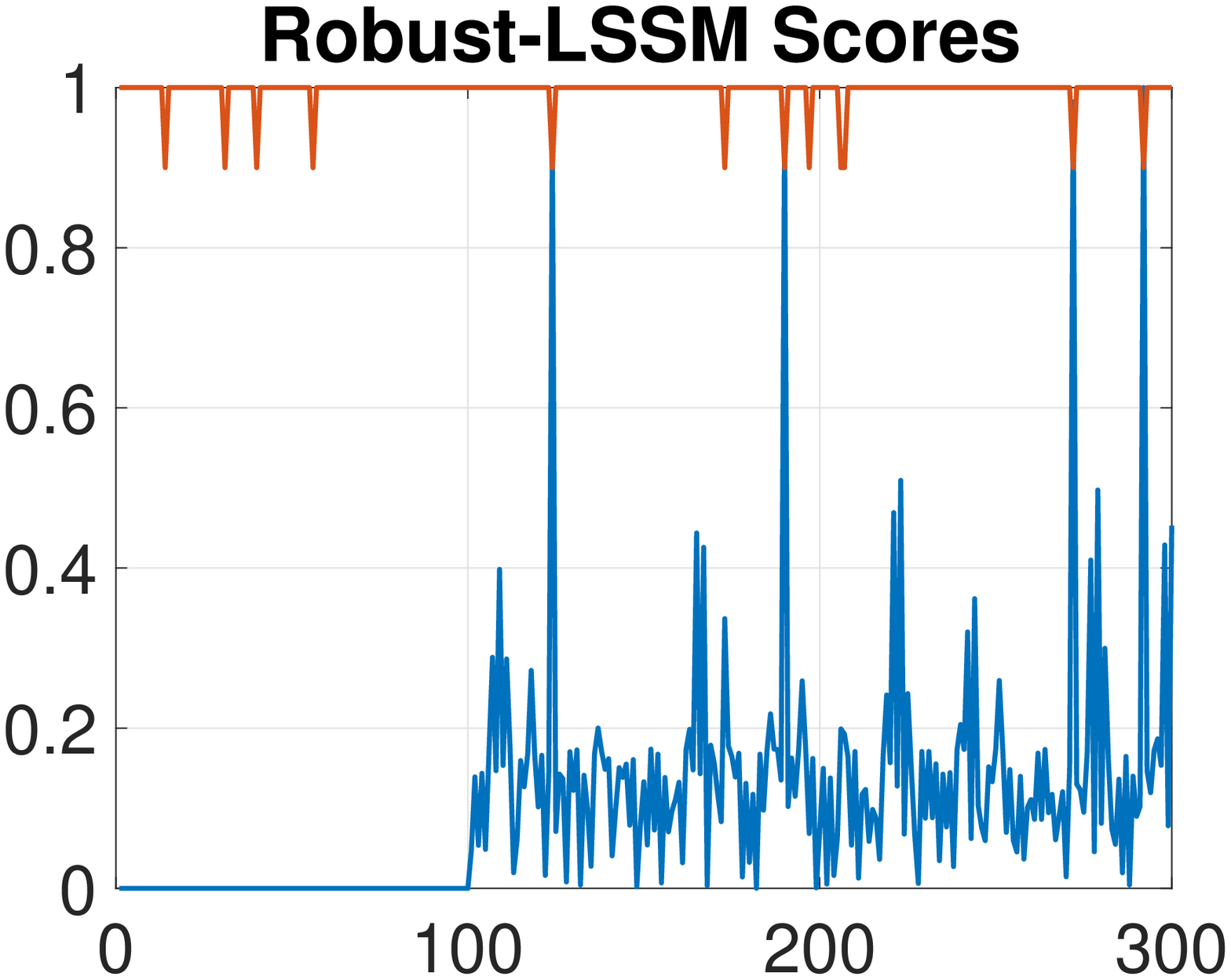}
}
\end{center}
%\vspace{-0.25in}
           \caption{This figure shows a time-series with several contextual anomalies as described in Section \ref{sec:versus_amp}. The red lines in each plot indicates the location of the outlying time-stamps. The plotted scores indicate that RPE successfully identifies even the weak anomalies. Each algorithm used the first 100 time-stamps as training data (refer to Section \ref{sec:versus_amp} for further details). }    
           \label{fig:contextual}\vspace{-0.18in}
\end{figure*}

%\vspace{-0.1in}
\section{Numerical Experiments}
%\vspace{-0.1in}
In this section, we compare the performance of the proposed approach against popular alternatives applicable in low data regimes where we have a limited number of samples from a time-series for training (in all the presented experiments, we provide no more than 100 training time-stamps). The methods that we compare against include Random Cut Forest (RCF) \cite{guha2016robust}, Robust Linear State Space Model (LSSM) \cite{durbin2012time}, Spectral Residual (SR) \cite{ren2019time}, IID, Autoregressive model \cite{chandola2009anomaly},  and Simple Projection based anomaly Extraction (SPE). The size of the window is set equal to 30 in all the experiments. %For AR, RPE, and SPE, we employed similar retraining schedules. 

In the RCF method, we set the number of trees equal to 30, and the sample size equal to 256. 
In the IID method, a memory of 100 last observed values is used to compute the mean and the variance which are then used to compute the p-value of each observation value based on Gaussian distribution. The anomaly score is equal to $1 - p$ where $p$ is the p-value. The Robust LSSM algorithm is a Linear Gaussian State Space Model equipped with dynamic masking where the algorithm considers the observations whose likelihood values are small as missing values and those observations are not used to perform filtering in the Kalman Filter.  Specifically, we compute the empirical CDF value of each generated score by the LSSM model using the memory of previously generated scores and those observations whose empirical CDF values are larger than 0.95 were not used to filter the state of the model. This technique helps the model to generate sharp anomaly scores.  
The used parameters for the RPE algorithm in all the presented experiments are $M_1 = 30$,  $q=100$, $t_{\max} = 300$, and $n_s = 5$.

The SPE algorithm is similar to the proposed RPE algorithm with only one difference. The difference is that SPE employs a simple projection method to compute the residual values corresponding to each observation vector. 
The SR algorithm requires computing Fast Fourier Transform (FFT) and inverse FFT for each time-stamp to compute the residual value. The length of the widow used for the computation of FFT  was chosen equal to 128. %Smaller window sizes degraded the performance of the algorithm and larger windows did not improve the performance of the algorithm. 

\begin{remark}
We set $r$ to equal the number of the singular values of $\bX\bX^T$ which are greater than $s_1 / 100$ where $s_1$ is the largest singular value of $\bX\bX^T$. If the estimated rank is larger than 10 (which can happen for random time-series that do not contain a background temporal pattern), we set $r=10$.
\end{remark}

\begin{remark}
%In many applications/businesses, it is not feasible to collect a large amount of data which suffices to train a deep learning based method. In addition, in most cases, we  do not have more than a few time-series which are similar/related to each other.
In  the experiments below all the models are trained locally, a separate model is trained for each time-series. This is a more realistic setting for many applications where we do not have a large number of related time-series.
For instance, if a dataset contains 5 time-series and the AR algorithm is used, the AR algorithm learns a different weight vector for each time-series. 
\end{remark}

\noindent
\textbf{Ablation study:} The key component of the proposed method which distinguishes it from the Singular spectrum analysis based algorithms is the robust projection step and its closed form implementation. In order to show its significance, we included the performance of SPE in all the presented experiments. The performance of SPE shows the performance of RPE in absence of the robust projection step.

\noindent
\textbf{ Probabilistic scores:} In some applications, the anomaly detection algorithm is required to provide  probabilistic anomaly scores (e.g, one minus p-value). Similar to most other anomaly detection algorithms (e.g., RCF, SR), RPE does not generate probabilistic scores and its outputs are the residual values. RPE transforms the given time-series with a possibly complicated background pattern into a simple series of residual values where the anomalies appear as spikes. A possible approach to transform the residual values into probabilistic scores is to learn the distribution of the residual values. For instance, the algorithm proposed in \cite{siffer2017anomaly} which employs extreme value theory can be used to infer the distribution of the residual values. Another approach is to apply an anomaly detection algorithm which generates probabilistic scores (e.g., LSSM) to the generated residual values.  In all the presented experiments, we report the results for a model called LRPE which is the combination of RPE and LSSM, i.e., LSSM is applied to the output of the RPE algorithm. Since we could not fit  the results of all the models in a table, we report the results of LRPE in each section inside the text. 

%The appendix includes further numerical investigations and visual illustration of the time-series in the presented experiments. 

 %   f1 = int(np.random.choice(list(range(40, 70)), 1))
    %f2 = int(np.random.choice(list(range(20, 40)), 1))
    %f3 = int(np.random.choice(list(range(10, 18)), 1))
    % f4 = int(np.random.choice(list(range(2, 6)), 1))

%\vspace{-0.1in}

\smallbreak

\subsection{Performance analysis with synthetic data}  
%\vspace{-0.1in}
In this experiment, we use synthetic data with a  seasonal pattern to evaluate the performance of the algorithms. All the time-series used in this experiment are generated as 
$$
\bt(j) =  \sum_{k=1}^4 z_k \cos \left(2 \pi \omega_k j + \psi_k \right) + \bn(j) \:.
$$
For each time-series $1/\omega_1$ is sampled randomly from $(40, 70)$, $1/\omega_2$ is sampled randomly from $(20, 40)$, $1/\omega_3$ is sampled randomly from $(10, 20)$, and $1/\omega_4$ is sampled randomly from $(2, 6)$. The phases $\{ \psi_k \}_{k=1}^4$ are sampled randomly from $(0, 2\pi)$. We assigned higher weights to the components with lower frequency. Specifically, we used $[z_1, z_2, z_3, z_4] = [2, 1.6, 1.2, 0.8]$.  
The first two (from right) time-series of Figure \ref{fig:ranks} shows two examples of the generated time-series. 
In all the experiments with synthetic data, 4 $\%$ of the time-stamps of each time-series are anomalies and results are obtained as the average of 20 independent runs. In each run, we obtain  max-$F_1$ accuracy along with its corresponding precision and recall. In all the tables, $F_1$, P, and R refer to  max-$F_1$, precision, and recall, respectively.  

%\vspace{-0.1in}
\subsubsection{Detecting Contextual Anomalies}
\label{sec:versus_amp}
%\vspace{-0.1in}
In this experiment, we examine the ability of the algorithms in detecting anomalies whose amplitude such that they do not necessarily push the time-stamp value out of the  range of normal values. Table \ref{tab:versus f 1} and Table \ref{tab:versus f 2} show the performance of the algorithms for two scenarios  where the amplitude of the added point anomalies are equal to $f/2$ and $f$ where $f = (\beta_{0.9} - \beta_{0.1})$ and $\beta_{q}$ is equal to $q^{th}$ quantile of $\bt$. Clearly, detecting the anomalies whose amplitude is smaller is more challenging and the algorithm should know the local structure precisely to be able to identify the potential deviations.  One can observe that in both cases, RPE yields the best performance which suggests that RPE is accurately aware of the local structure of the time-series. The obtained results for LRPE corresponding to the scenarios  in Table \ref{tab:versus f 1}  and Table \ref{tab:versus f 2} are ($F_1$=0.99, P$=$0.99, R$=$0.99) and ($F_1$=0.95, P$=$0.95, R$=$0.99), respectively. Figure \ref{fig:contextual} shows an example where the time-series contains several contextual anomalies. 

\begin{remark}
The reported $F_1$-score, precision, and recall are the average of $F_1$-score, precision, and recall of 20 independent runs. Therefore, the reported $F_1$-score is not necessarily equal to the harmonic mean of the reported precision and recall.
\end{remark}

 \begin{table}[h]
\begin{center}{
\caption{Accuracy of the algorithms. Anomaly amplitude equal to $f$ (Experiment \ref{sec:versus_amp}).
}
\label{tab:versus f 1}
\begin{tabular}{l |lcccccc}
\hline\hline
  &     RCF  &  RPE  & SPE  & IID  &  SR &  AR &  LSSM \\
  \hline
$F_1$  &     0.74  &  \textbf{1}  & 0.96  & 0.64  &  0.81 &  0.97 &  0.93 \\
      \hline
P  &     0.86  &  1  & 0.95  & 0.79 &   0.85 &  0.99 &  0.9 \\
      \hline
R  &     0.68  &  1  & 0.96  & 0.58  &  0.80 &  0.96&  0.98 \\
 \hline\hline
\end{tabular}
}
\end{center}
\vspace{-0.2in}
\end{table}

 \begin{table}[h]
\begin{center}{
\caption{Accuracy of the algorithms. Anomaly amplitude equal to $f/2$ (Experiment \ref{sec:versus_amp}).
}
\label{tab:versus f 2}
\begin{tabular}{l |lcccccc}
\hline\hline
  &     RCF  &  RPE  & SPE  & IID  &  SR &  AR &  LSSM \\
  \hline
$F_1$  &     0.42  &  \textbf{0.96}  & 0.92  & 0.25  &  0.38 &  0.86 &  0.55 \\
      \hline
P  &     0.52  &  0.96  & 0.92  & 0.36 &  0.54 &  0.86 &  0.63 \\
      \hline
R  &     0.43  &  0.98  & 0.92  & 0.33 &  0.37 &  0.89 &  0.60 \\
 \hline\hline
\end{tabular}
}
\end{center}
\vspace{-0.2in}
\end{table}

\subsubsection{Identifying range anomalies}
\label{sec:range_anomaly}
In this experiment, we investigate the performance of the models when the length of each anomaly is more than a single time-stamp.  Table \ref{tab:2timestamps} shows the performance of the models  when each anomaly corrupts 2 consecutive time-stamps and Table \ref{tab:4timestamps} shows the results when   each anomaly corrupts 4 consecutive time-stamps. In both cases, the amplitude of the added anomalies is equal to $f/1.5$
and the standard deviation of  the added noise is equal to 0.1. One can observe that the performance of all the algorithms degrade when the range of anomalies increases in time domain and in both cases RPE yields the best accuracy. The obtained results for LRPE corresponding to the scenarios in Table \ref{tab:2timestamps}  and Table \ref{tab:4timestamps} are ($F_1$=0.94, P$=$0.98, R$=$0.91) and ($F_1$=0.83, P$=$0.82, R$=$0.87), respectively.

 \begin{table}[h]
\begin{center}{
\caption{The length of each anomaly equal to 2 time-stamps.  (Experiment \ref{sec:range_anomaly}).
}
\label{tab:2timestamps}
\begin{tabular}{l |lcccccc}
\hline\hline
  &     RCF  &  RPE  & SPE  & IID  &  SR &  AR &  LSSM \\
  \hline
$F_1$  &     0.5  &  \textbf{0.97}  & 0.77  & 0.40 &  0.43 &  0.93 &  0.66 \\
      \hline
P  &     0.67  &  0.98  & 0.85  & 0.46 &   0.63 &  0.96 &  0.68 \\
      \hline
R  &     0.48  &  0.96  & 0.74  & 0.49 &   0.4 &  0.91 &  0.67  \\
 \hline\hline
\end{tabular}
}
\end{center}
\vspace{-0.2in}
\end{table}

 \begin{table}[h]
\begin{center}{
\caption{The length of each anomaly equal to 4 time-stamps.   (Experiment \ref{sec:range_anomaly}).
}
\label{tab:4timestamps}
\begin{tabular}{l |lccccccc}
\hline\hline
  &     RCF  &  RPE  & SPE  & IID  &  SR &  AR &  LSSM \\
  \hline
$F_1$  &     0.51    &  \textbf{0.83}  & 0.55  & 0.42 &  0.33 &  0.76 &  0.5 \\
      \hline
P  &     0.59  &  0.81  & 0.59  & 0.44  &  0.52 &  0.80 &  0.63 \\
      \hline
R  &     0.63  &  0.87  & 0.59  & 0.68  &  0.43 &  0.79 &  0.51  \\
 \hline\hline
\end{tabular}
}
\end{center}
\vspace{-0.2in}
\end{table}

 \begin{table}[h]
\begin{center}{
\caption{Accuracy of the algorithms. Real time-series sampled from Yahoo dataset (Experiment \ref{sec:yahoo}).
}
\label{tab:yahoo}
\begin{tabular}{l |lccccccc}
\hline\hline
  &     RCF  &  RPE  & SPE  & IID &  SR &  AR &  LSSM \\
  \hline
$F_1$  &     0.72  &  \textbf{0.88}  & 0.77  & 0.72 &  0.72 &  0.79 &  0.84 \\
      \hline
P  &      0.77  &  0.92  & 0.87  & 0.88  &  0.86 &  0.89 &  0.85 \\
      \hline
R  &      0.72  &  0.88  & 0.74  & 0.66  &  0.67 &  0.74 &  0.87 \\
 \hline\hline
\end{tabular}
}
\end{center}
\vspace{-0.2in}
\end{table}

\subsection{Experiment with real data: Yahoo}
\label{sec:yahoo}
In this experiment, we use the real time-series in the Yahoo dataset. Yahoo dataset is an open data  for anomaly detection released by Yahoo. The real time-series in the Yahoo dataset are long (mostly around 1400 time-stamps) and in order to create challenging/realistic scenarios, we sample random sub-sequences from each time-series with length 300 where the first 100 time-stamps were used as training data (if the frequency is 1 day, one needs to wait for about 4 years to collect 1400 samples!). Specifically, for each real time-series, we sampled 15 random sub-time-series with length equal to 300. Since these sub-time-series might not include any anomaly, we added anomalies to each sampled time-series to ensure that 4 $\%$ of the time-stamps are anomalies (we used similar technique used for the synthetic data to add anomalies to the real time-series and half of the added anomalies were contextual anomalies). 
Table \ref{tab:yahoo} shows the accuracy of different models and the obtained results for LRPE  are ($F_1$=0.86, P$=$0.93, R$=$0.83). One can observe that RPE yields the best accuracy among the models. The main reason is that the robust projection step enables the RPE algorithm to robustly compute the residual values even if its window contains corrupted time-stamps. In addition, it enables the algorithm to distinguish the outlying time-stamps even if they are close to each. \\
\textbf{Remark.} Similar experiments with two more real dataset are included in the appendix.

%\subsection{}
\subsection{Choosing window size $M_1$}
\label{sec:choosM1}
The main idea behind the RPE algorithm was to utilize the low dimensional structure of the trajectory matrix $\bX \in \mathbb{R}^{M_1 \times M_2}$ to robustly extract the outlying components using the column space of $\bX$. Therefore, $M_1$ should be sufficiently large such that $\bX$ is a low rank matrix. The presented theoretical results also emphasised that $r/M_1$ should be sufficiently small to guarantee the performance of the robust projection step because the value of $\mu(\bU)$ increases in most cases when $r/M_1$ increases (for instance, for a randomly generated subspace $\bU \in \mathbb{R}^{M_1 \times r}$, $\mathbb{E}(\| \be_i^T \bU \|_2^2) = \frac{r}{M_1}$ \cite{candes2009exact}).
In  the real and synthetic time-series that we studied, the ranks of the time-series which contained temporal background signal were smaller than 6 in most of the cases. Therefore, we chose $M_1 = 30$ in the presented experiments which means that the dimension of the matrix is at least 5 times larger than the rank.
If there is not a known upper-bound for $r$, one can form multiple trajectory matrices  with different values of $M_1$ and find the proper value which ensures that $\bX$ is a low rank matrix. 

Although $M_1$ should be sufficiently large to ensure that $r/M_1$ is  small enough,  choosing $M_1$ too large also  hurt the performance of the model because $M_2 = n - 1 - M_1$ where $M_2$ is the number of columns of $\bX$ and a large window size makes $M_2$ small. Note that in order to ensure that $\bX$ is a low rank matrix, both $M_1$ and $M_2$ should be sufficiently larger than $r$. Moreover, choosing a very large value for $M_1$ can limit the understanding of the model from the local structures of the time-series. Our investigations suggest that choosing $5r \le M_1 \le 10 r$ works best. For instance, Table \ref{tab:chooseM1} shows the performance of RPE for a scenario similar to the experiment in Section \ref{sec:range_anomaly} (length of range anomalies equal to 2) where the length of the training time-series is equal to 100. The results show that choosing a too small or a too large value for $M_1$ degrade the performance and the performance is not sensitive to the value of $M_1$ as long as we choose a reasonable value. 
%In addition, one can add anomalies to  the train time-series to choose the best value of $M_1$ which ensures the low-rankness of $\bX$ and achieves the best detection accuracy over the training data. In the presented experiments, we did not optimize the value of $M_1$ and set it equal to 30 for all the experiments. 
%We did not find the performance of the proposed algorithm  to be sensitive to the value of $M_1$.

 \begin{table}[h]
\begin{center}{
\caption{The accuracy of RPE with different values of $M_1$   (Experiment \ref{sec:choosM1}).
}
\label{tab:chooseM1}
\begin{tabular}{l |lccccc}
\hline\hline
 $M_1$ &     30  &  40  & 50  & 10  &  90 &   \\
  \hline
  \\
 $F_1$ &     0.97  &  0.98  & 0.98  & 0.42  &  0.68 &   \\
  \hline
 \hline
\end{tabular}
}
\end{center}
\vspace{-0.2in}
\end{table}

\section{Conclusion}
A novel, closed-form, and efficient anomaly detection algorithm for univariate time-series were proposed. The proposed method, dubbed RPE, leverages the linear structure of the trajectory matrix and it employs a  robust projection method to identify the corrupted time-stamps. A provable and closed-form algorithm was presented which enables the algorithm to perform the robust projection step without the need to run an iterative solver for each time-stamp. The presented experiments showed that RPE can outperform the existing approaches with a notable margin.

\bibliography{example_paper}
\bibliographystyle{apalike}

\newpage

%\subsection{Estimating $\calU$}

\subsection*{Proofs of the theoretical results}
\textbf{Proof of Lemma \ref{lm:my_candes}}\\
In order to show that $\ba$ is the optimal point of (\ref{eq:elaa}), it is enough to show that 
 \begin{eqnarray}
  \begin{aligned}
 \| \bx - \bU ({\ba} + \bh) \|_1 -  \| \bx - \bU {\ba}  \|_1 > 0 
  \end{aligned}
  \label{eq:22}
\end{eqnarray}
for any sufficiently small  non-zero perturbation vector $\bh$.  Note that $\bx = \bU \ba + \bs$ which means that (\ref{eq:22}) can be simplified as 
 \begin{eqnarray}
  \begin{aligned}
\sum_{i \in \calI_s} \left| \bs(i) - \bu_i^T \bh  \right| + \sum_{i \notin \calI_s} \left| \bu_i^T \bh \right| - \sum_{i \in \calI_s}  \left| \bs(i)  \right| > 0 \:,
  \end{aligned}
  \label{eq:33}
\end{eqnarray}
where $\calI_s$ is the set of the indices of the non-zero elements of $\bs$.
According to (\ref{eq:33}), it suffices to ensure that
 \begin{eqnarray}
  \begin{aligned}
\sum_{i \notin \calI_s} \left| \bu_i^T \bh \right| - \sum_{i \in \calI_s} \left|  \bu_i^T \bh  \right|   > 0 \:,
  \end{aligned}
  \label{eq:44}
\end{eqnarray}
which is equivalent to guaranteeing that
 \begin{eqnarray}
  \begin{aligned}
 \| \bU \bh \|_1 - 2 \sum_{i \in \calI_s} \left|  \bu_i^T \bh  \right|   > 0 \:.
  \end{aligned}
  \label{eq:55}
\end{eqnarray}
Since $\bh$ appears on both sides of (\ref{eq:55}), it is enough to ensure that 
 \begin{eqnarray}
  \begin{aligned}
 \min_{\bh \in \mathbb{S}^{r-1}} \| \bU \bh \|_1 - 2 \sum_{i \in \calI_s} \| \bu_i \|_2  > 0 \:.
  \end{aligned}
  \label{eq:56}
\end{eqnarray}
Therefore, according to the definition of  $\mu$$$\mu^2(\bU) = \max_i \frac{\| \be_i ^T \bU \|_2^2}{r} \:,$$
where $\be_i$ is the $i^{th}$ column/row of the identity matrix,
it is enough to ensure that 
 \begin{eqnarray}
  \begin{aligned}
 \min_{\bh \in \mathbb{S}^{r-1}} \| \bU \bh \|_1 - 2 m \sqrt{r} \mu (\bU)> 0 \:.
  \end{aligned}
  \label{eq:57}
\end{eqnarray}
Thus, if 
$
\kappa (\bU)< \frac{1}{2 m \: \sqrt{r}}\:,
$
then (\ref{eq:22}) is guaranteed to hold which means that $\ba$ is the optimal point of (\ref{eq:elaa}) where $\kappa (\bU) = {\mu (\bU)}{\gamma(\bU)}$ and $\gamma(\bU) = \frac{1}{\min_{\bh \in \mathbb{S}^{r-1}} \| \bU \bh \|_1}$.

\noindent
\textbf{Proof of Lemma \ref{lm:closed-form}}\\
The vector of residual values $\hat{\be} = |\bx - \bU \bU^T \bx|$ can be written as  $\hat{\be} = | \bU^{\perp} {\bU^\perp}^T \bx|$ where $\bU^{\perp} \in \mathbb{R}^{M_1 \times (M_1 - r)}$ is an orthonormal basis for the complement of the span of $\bU$.  Define $\calI_s$ as the set of the indices of the non-zero elements of $\bs$. The vector $\hat{\be} \in \mathbb{R}^{M_1}$ can be written as
 \begin{eqnarray}
  \begin{aligned}
\hat{\be} = \left| \bU^{\perp}  \sum_{j \in \calI_s} \bs(j) \bu^{\perp}_{j}  \right| \:,
  \end{aligned}
    \label{eq:e1}
\end{eqnarray}
where $\bs(j)$ is the $j^{th}$ element of $\bs$ and $\bu^{\perp}_{j}$ is the $j^{th}$ row of $ \bU^{\perp}$. Therefore, the $i^{th}$ element of  $\hat{\be}$ can be described as
 \begin{eqnarray}
  \begin{aligned}
\hat{\be}(i) = \left| \sum_{j \in \calI_s} \bs(j) {\bu^{\perp}_i}^T \bu^{\perp}_{j}  \right| \:.
  \end{aligned}
\end{eqnarray}

Since $ \bU^{\perp}$ is the span of the complement of the span of $\bU$ and they both are orthonormal matrices, we can conclude that 
 \begin{eqnarray}
  \begin{aligned}
& {\bu^{\perp}_i}^T \bu^{\perp}_{j}  = - {\bu_i}^T \bu_{j} \quad \text{if} \quad i \neq j \\
& {\bu^{\perp}_i}^T \bu^{\perp}_{i}  = 1 - {\bu_i}^T \bu_{i} 
  \end{aligned}
  \label{eq_eq}
\end{eqnarray}
where $\bu_i$ is the $i^{th}$ row of $\bU$.
Now we consider different scenarios for $\hat{\be}(i)$.
When $i \notin     \calI_s$
 \begin{eqnarray}
  \begin{aligned}
 \hat{\be}(i) & = \left| \sum_{j \in \calI_s} \bs(j) {\bu^{\perp}_i}^T \bu^{\perp}_{j}  \right|
=  \left| - \sum_{j \in \calI_s} \bs(j) {\bu_i}^T \bu_{j}  \right| \:,
  \end{aligned}
\end{eqnarray}
and when $i \in     \calI_s$
 \begin{eqnarray}
  \begin{aligned}
 \hat{\be}(i) & = \left| \sum_{j \in \calI_s} \bs(j) {\bu^{\perp}_i}^T \bu^{\perp}_{j}  \right|
=  \left| \bs(i) -  \bs(i) \bu_i^T \bu_i - \sum_{j \in \calI_s \atop j \neq i} \bs(j) {\bu_i}^T \bu_{j}  \right| \:.
  \end{aligned}
  \label{eq:with_minus}
\end{eqnarray}
According to the definition of $\mu(\bU)$,
 $$\bu_i^T \bu_j \leq r\:\mu^2(\bU) \:.$$
Therefore, when $i \notin \calI_s$, we can set the following upper-bound for $ \hat{\be}(i) $
 \begin{eqnarray}
  \begin{aligned}
 \hat{\be}(i) & =  \left| - \sum_{j \in \calI_s} \bs(j) {\bu_i}^T \bu_{j}  \right|  \leq r \: \mu^2(\bU)  \sum_{j \in \calI_s} \left|\bs(j) \right|
  \end{aligned}
    \label{eq: upper_no_noise}
\end{eqnarray}
and the following lower-bound for $i \in \calI_s$ 
 \begin{eqnarray}
  \begin{aligned}
\hat{\be}(i) & =\left| \bs(i) -  \bs(i) \bu_i^T \bu_i - \sum_{j \in \calI_s \atop j \neq i} \bs(j) {\bu_i}^T \bu_{j}  \right| \\
& \ge |\bs(i)| -  \sum_{j \in \calI_s} \left| \bs(i) \bu_i^T \bu_j \right| \ge  |\bs(i)| - r \: \mu^2(\bU)  \sum_{j \in \calI_s} \left|\bs(j) \right| \:.
  \end{aligned}
  \label{eq: lower_no_noise}
\end{eqnarray}

According to (\ref{eq: upper_no_noise}) and (\ref{eq: lower_no_noise}),
in order to ensure that the algorithm samples all the corrupted time-stamps, it is enough to guarantee that 
  \begin{eqnarray}
  \begin{aligned}
\mu^2(\bU) \le \frac{1}{2 r} \frac{\min_{i} \zeta_i }{\sum_{j} \zeta_j} \:,
  \end{aligned}
\end{eqnarray}
where $\{ \zeta_i\}_{i=1}^m$ are the absolute value of the non-zero elements of $\bs$ and $m = |\calI_s|$.
\\
\\
\textbf{Proof of Theorem \ref{thm:withnoise}}
\\
In order to prove Theorem \ref{thm:withnoise}, first we derive the sufficient condition which guarantees that 
  \begin{eqnarray}
  \begin{aligned}
\min_{i \in \calI_s} \hat{\be}(i)  > \max_{i \notin \calI_s} \hat{\be} (i)\:.
  \end{aligned}
\end{eqnarray}

When the observation vector $\bx$ contains added noise, then similar to (\ref{eq:e1}), $\hat{\be}$ can be written as 
 \begin{eqnarray}
  \begin{aligned}
\hat{\be} = \left| \bU^{\perp}  \left(\sum_{j \in \calI_s} \bs(j) \bu^{\perp}_{j} + \sum_{k=1}^{M_1} \bn(k) \bu_k^{\perp} \right)  \right| \:.
  \end{aligned}
\end{eqnarray}
Thus, the $i^{th}$ element of $\hat{\be}$ can be expanded as follows
 \begin{eqnarray}
  \begin{aligned}
\hat{\be}(i) = \left| \sum_{j \in \calI_s} \bs(j) {\bu^{\perp}_i}^T \bu^{\perp}_{j}  + \sum_{k=1}^{M_1} \bn(k) {\bu^{\perp}_i}^T \bu^{\perp}_{j} \right| \:.
  \end{aligned}
\end{eqnarray}
Similar to the proof of Lemma \ref{lm:closed-form}, we set a lower bound for $\hat{\be}(i)$ when $i \in \calI_s$ and a upper-bound for $\hat{\be}(i)$ when 
when $i \notin \calI_s$. The upper-bound for the cases where $i \notin \calI_s$ can obtained as 
 \begin{eqnarray}
  \begin{aligned}
 \hat{\be}(i) & = \left| \sum_{j \in \calI_s} \bs(j) {\bu^{\perp}_i}^T \bu^{\perp}_{j}  + \sum_{k=1}^{M_1} \bn(k) {\bu^{\perp}_i}^T \bu^{\perp}_{j} \right| \\
 & \le  \sum_{j \in \calI_s} \left| \bs(j) {\bu^{\perp}_i}^T \bu^{\perp}_{j} \right| +\left| \sum_{k=1}^{M_1}  \bn(k) {\bu^{\perp}_i}^T \bu^{\perp}_{j} \right|
 \\
% & \leq r\: \mu^2(\bU)  \left( \sum_{j \in \calI_s} \left|\bs(j) \right| +  \sum_{k=1}^{M_1} |\bn(k)| \right) \:.
   \end{aligned}
   \label{eq:initial_up}
\end{eqnarray}

We did not specify any distribution for the added noise vector $\bn$ but we made the following two assumptions
 \begin{eqnarray}
  \begin{aligned}
& \mathbb{P}\left[ | \bn(k)| > \eta \right] < \delta / M_1\\
& \mathbb{P}\left[ | \bn(k)| > 0 \right] = \mathbb{P}\left[ | \bn(k)| < 0 \right] \:.
   \end{aligned}
\end{eqnarray}
In order to bound the second part of (\ref{eq:initial_up}), we utilize the Hoeffding's inequality \cite{wainwright2019high}.

\begin{lemma} 
Suppose that $\{ x_i\}_{i=1}^{m}$ are bounded random variables such that $a \leq x_i \leq b$ and assume that $\{ \epsilon_i \}_{1=1}^m$ are sampled independently from the Rademacher distribution. Then, 
 \begin{eqnarray}
  \begin{aligned}
\mathbb{P} \left[\left | \sum_{i=1}^m \epsilon_i x_i  \right| > t\right] < \exp\left(-\frac{2 t^2}{m (b - a)^2}\right) \:.
   \end{aligned}
\end{eqnarray}
\label{lem:hoeff}
\end{lemma}

Since $\mathbb{P}\left[ | \bn(k)| > 0 \right] = \mathbb{P}\left[ | \bn(k)| < 0 \right]$, the distribution of $$\left| \sum_{k=1}^{M_1}  \bn(k) {\bu^{\perp}_i}^T \bu^{\perp}_{j} \right|$$ is equivalent to the distribution of $\left| \sum_{k=1}^{M_1} \epsilon_i  \left| \bn(k) {\bu^{\perp}_i}^T \bu^{\perp}_{j} \right| \right|$ where $\{\epsilon_i \}_{i=1}^{M_1}$ are sampled independently from Rademacher distribution.
Therefore, using Lemma \ref{lem:hoeff} and conditioned on $\bn(k) < \eta$, we can conclude that
 \begin{eqnarray}
  \begin{aligned}
\mathbb{P} \left[ \left| \sum_{k=1}^{M_1}  \bn(k) {\bu^{\perp}_i}^T \bu^{\perp}_{j} \right| > t \right]\leq \exp\left( - \frac{2 t^2}{M_1 \mu^4 (\bU) r^2 \eta^2} \right)  \:,
   \end{aligned}
   \label{eq:tt1}
\end{eqnarray}
where we used the fact that $|{\bu^{\perp}_i}^T \bu^{\perp}_{j} | = |{\bu^{}_i}^T \bu^{}_{j} |$ when $i \neq j$.
Thus, according to (\ref{eq:tt1}) and the presumed model for noise, 
 \begin{eqnarray}
  \begin{aligned}
\mathbb{P} \left[ \left| \sum_{k=1}^{M_1}  \bn(k) {\bu^{\perp}_i}^T \bu^{\perp}_{j} \right| > r \mu^2(\bU) \eta \sqrt{\frac{M_1 \log 1/ \delta}{2}}  \right]\leq 2 \delta \: .
   \end{aligned}
   \label{eq:boundnoise}
\end{eqnarray}
Using (\ref{eq:initial_up}) and (\ref{eq:boundnoise}) we can establish the following upper-bound for $\hat{\be(i)}$ for the cases where $i \notin \calI_s$
 \begin{eqnarray}
  \begin{aligned}
\mathbb{P}\Bigg[ \hat{\be}(i) < r \mu^2 (\bU) \sum_{j=1}^m \zeta_j + r \mu^2(\bU)  \eta \sqrt{\frac{M_1 \log 1/ \delta}{2}} \Bigg] \geq 1 - 2 \delta  \:,
   \end{aligned}
     \label{eq:f_upper_bound}
\end{eqnarray}
where $\{ \zeta_i\}_{i=1}^m$ are the absolute value of the non-zero elements of $\bs$ and $m = |\calI_s|$.
When $i \in \calI_s$, similar to (\ref{eq:with_minus}), we can rewrite $\hat{\be(i)}$ as
 \begin{eqnarray}
  \begin{aligned}
  \hat{\be}(i) & = \Bigg| \bs(i) -  \bs(i) \bu_i^T \bu_i - \sum_{j \in \calI_s \atop j \neq i} \bs(j) {\bu_i}^T \bu_{j}    - \sum_{k} \bn(k) {\bu^{}_i}^T \bu^{}_{j}  \Bigg| \\
 &  \geq \left| \bs(i) \right| - \left| \sum_{j \in \calI_s} \bs(j) {\bu_i}^T \bu_{j}   \right|  - \left| \sum_{k} \bn(k) {\bu^{}_i}^T \bu^{}_{j}  \right| \:.
  \end{aligned}
  \label{eq:lower1}
\end{eqnarray}
Using similar techniques used to establish (\ref{eq:f_upper_bound}) and according to (\ref{eq:lower1}), 
we can conclude that 
 \begin{eqnarray}
  \begin{aligned}
\hat{\be}(i) & \geq \left| \bs(i) \right| - \sum_{j \in \calI_s}  \left| \bs(j) {\bu_i}^T \bu_{j}   \right|  - \left| \sum_{k} \bn(k) {\bu^{}_i}^T \bu^{}_{j}  \right| \\
&\geq  \left| \bs(i) \right| - r \mu^2(\bU)  \sum_j \zeta_j - r \mu^2(\bU)  \eta \sqrt{\frac{M_1 \log 1/ \delta}{2}}
  \end{aligned}  
  \label{eq:to2}
\end{eqnarray}
with probability at least $1 - 2\delta$. Thus, according to (\ref{eq:f_upper_bound}) and (\ref{eq:to2}), we can conclude that if 
 \begin{eqnarray}
  \begin{aligned}
\mu^2(\bU)  \left(1 + \frac{\eta}{\sum_j \zeta_j } \sqrt{\frac{M_1 \log 1/ \delta}{2}} \right) \le \frac{1}{2 r} \frac{ \min_j \zeta_j}{\sum_j \zeta_j} \:,
  \end{aligned}  
\end{eqnarray}
then all the indices of the non-zero elements of $\bs$ are successfully rejected by the algorithm. 
%The parameter $\eta$ as was defined as $\mathbb{P}\left[ | \bn(i) | > \eta\right] \le \frac{\eta}{M_1}$. Therefore, (\ref{eq:initial_up}) can be simplified to 
% \begin{eqnarray}
  % \begin{aligned}
 %\mathbb{P} \left[\hat{\be}(i) \leq r\: \mu^2(\bU)  \left( \sum_{j \in \calI_s} \left|\bs(j) \right| + M_1 \eta \right) \right] > 1 - \delta \:.
%   \end{aligned}
% \end{eqnarray}

\end{document}